\documentclass[journal]{IEEEtran}
\ifCLASSINFOpdf
\else
\fi
\usepackage[dvipsnames]{xcolor}
\usepackage{tikz}
\usepackage{amsmath}
\usepackage{tabularx}
\usepackage{bm}
\usepackage{diagbox}
\usepackage{multirow}
\usepackage{float}
\usepackage{tfrupee}
\usepackage{enumitem}
\usepackage{subfig}
\usepackage{graphicx}
\usepackage{comment}
\usepackage{amssymb}
\usepackage{url}
\usepackage{hyperref}
\usepackage[linesnumbered,ruled,vlined]{algorithm2e}
\SetKwInput{KwInput}{Input}                
\SetKwInput{KwOutput}{Output}
\SetAlFnt{\small}

\DeclareMathOperator*{\position}{index}
\newcommand{\chen}[1]{\textcolor{red}{\textbf{Chen}: #1}}

\begin{document}
%
\title{Smart App Attack: Hacking Deep Learning Models in Android Apps}
%
%
%

\author{Yujin Huang,
        Chunyang Chen*
\thanks{Y. Huang is with the Faculty of Information Technology, Monash University, Melbourne, Victoria 3800, Australia (e-mail: yujin.huang@monash.edu).}
\thanks{C. Chen is with the Faculty of Information Technology, Monash University, Melbourne, Victoria 3800, Australia (e-mail: chunyang.chen@monash.edu).}
\thanks{*corresponding author}}

\maketitle
\begin{abstract}
On-device deep learning is rapidly gaining popularity in mobile applications.
Compared to offloading deep learning from smartphones to the cloud, on-device deep learning enables offline model inference while preserving user privacy.
However, such mechanisms inevitably store models on users' smartphones and may invite adversarial attacks as they are accessible to attackers.
Due to the characteristic of the on-device model, most existing adversarial attacks cannot be directly applied for on-device models.
In this paper, we introduce a grey-box adversarial attack framework to hack on-device models by crafting highly similar binary classification models based on identified transfer learning approaches and pre-trained models from TensorFlow Hub. 
We evaluate the attack effectiveness and generality in terms of four different settings including pre-trained models, datasets, transfer learning approaches and adversarial attack algorithms.
The results demonstrate that the proposed attacks remain effective  regardless of different settings, and significantly outperform state-of-the-art baselines.
We further conduct an empirical study on real-world deep learning mobile apps collected from Google Play. 
Among 53 apps adopting transfer learning, we find that 71.7\% of them can be successfully attacked, which includes popular ones in medicine, automation, and finance categories with critical usage scenarios.
The results call for the awareness and actions of deep learning mobile app developers to secure the on-device models.
The code of this work is available at \href{https://github.com/Jinxhy/SmartAppAttack}{\textcolor{RubineRed}{\normalfont \textit{https://github.com/Jinxhy/SmartAppAttack}}}.
\end{abstract}

\begin{IEEEkeywords}
Deep learning, adversarial attack, mobile application, security.
\end{IEEEkeywords}

%
\IEEEpeerreviewmaketitle

\section{Introduction}
\label{sec:intro}

\IEEEPARstart{D}{eep} learning has already been proven powerful in several fields, including object detection, natural language question answering, and speech recognition.
Meanwhile, mobile apps are making people's life much convenient including reading, chatting, banking, etc.
To make mobile apps really "smart", many development teams have been adopting deep learning to equip their apps with artificial intelligence features, such as image classification, face recognition, and voice assistant.
There are two ways to host deep learning models for mobile apps, i.e., on cloud or on device within the app.

Compared to performing deep learning on cloud, on-device deep learning offers several unique benefits as follows.
First, it avoids sending private users' data to the cloud, leading to the bandwidth saving, inference accelerating, and privacy preserving~\cite{huang2021robustness}.
Second, apps can run in any situation with no need for internet connectivity~\cite{sun2021mind}.
Also considering the rapidly increasing computing power of mobile phones, more and more deep learning models are directly deployed to end-users' devices~\cite{xu2019first}.
On account of the benefits of embedding deep learning into mobile apps, mainstream deep learning frameworks also roll out their corresponding on-device versions such as TensorFlow Lite (TFLite)~\cite{tensorflowlite} and Pytorch Mobile~\cite{pytorchmobile}.

However, emerging adversarial attacks have exposed great threats to neural networks \cite{goodfellow2014explaining}, which alters input instances in the form of small perturbations that remain imperceptible to the human, but fool neural network classifiers in making incorrect predictions.
There are many different types of adversarial attacks including white-box (e.g., Fast Gradient Sign Method~\cite{suya2020hybrid} and Carlini and Wagner~\cite{chen2017zoo} attacks) and black-box ones (e.g., Zeroth Order Optimization~\cite{chen2020hopskipjumpattack} and Bandits and Priors~\cite{co2019procedural} attacks).
White-box attacks require attackers to have full access to the targeted model, whereas black-box attacks require only queries to the targeted model that may return complete or limited information.
In general, black-box attacks are more challenging and expensive compared to white-box ones.
Most adversarial attacks are targeting at on-cloud deep learning models~\cite{li2019adversarial, brendel2017decision}, while very few of them can be applied for attacking on-device models.
In view of the fact that many mobile apps supported by deep learning models are used for vital tasks such as financial, social or even life-critical tasks like medical diagnosis, driving assistant, face recognition, attacking the models inside those apps will post a severe threat to end-users.

Although there are some works investigating the on-device model attack or defense including stealing model~\cite{sun2021mind}, adding backdoor~\cite{li2021deeppayload} or testing model robustness~\cite{huang2021robustness}, most of them are empirical studies and merely provide universal methods to attack on-device models without taking the specific model characteristics (i.e., model structure and parameters) into account.
This leads to poor attack performance when on-device models adopt different model structures and tuning strategies.
In contrast, we are proposing a simple but effective attack to on-device models by taking their offline characteristics into consideration.   
Different from cloud-based deep learning models, hackers can obtain the deep learning model with explicit structural and parametric information.
But note that it is not fully white-box, as creators of on-device model frameworks noticed that security concern and take actions to prevent that leakage. 
For example, TFLite (the most popular on-device framework~\cite{huang2021robustness,xu2019first}) model is specified as irreversible to its trainable format in TensorFlow\footnote{Previous official reverse tool \url{https://github.com/tensorflow/tensorflow/tree/r1.9/tensorflow/contrib/lite} has been obsolete since TensorFlow 1.9}.
That is why we cannot directly apply existing white-box adversarial attacks.
Although we can enumerate a list of rules for manual conversion, it is not general and easy to be out of date as there may be customized operations defined by developers and multiple deep learning frameworks~\cite{wang2020edge}.
Hence, the attack to the on-device model can be regarded as a grey-box attack and we propose a way to bypass that limitation so that a white-box attack can be feasible.

In this paper, we propose a grey-box adversarial attack framework, which is designed specifically to hack the on-device models that adopt transfer learning.
Given a mobile app, we first extract the deep learning model inside by reverse-engineering the app.
We then check if the model is a fine-tuned one based on existing pre-trained models on public repositories (e.g., Tensorflow Hub~\cite{TensorHub}, Model Zoo~\cite{modelzoo}).
By identifying its pre-trained model and corresponding transfer learning approach, we obtain a trainable and similar white-box model.
For a specific targeted class, we then fine-tune a binary adversarial model that shares similar features (i.e., the features associated with the targeted class) with the victim on-device model.
Finally, we craft adversarial examples based on the binary adversarial model using existing white-box adversarial attack approaches to perform attacks.

To evaluate the attack effectiveness and generality of our framework, we conduct experiments with different settings, including pre-trained models, datasets, transfer learning approaches and adversarial attack algorithms.
These experiments independently test the impact of each perspective on attack performance.
The results show that our attacks remain effective regardless of different settings, 
with more than 2 times the success rate compared with the previous attack \cite{huang2021robustness}.
To further examine the feasibility of our framework on real-world mobile apps, we launch the attack on 53 deep learning mobile apps that adopt transfer learning.
Of these apps, 71.7\% can be successfully attacked, which includes popular ones in medicine, automation, and finance categories with critical usage scenarios such as skin cancer recognition, traffic sign detection, etc.

We issue this new attack to raise the awareness of the community for protecting the on-device models.
This paper makes the following research contributions:

\begin{itemize}[leftmargin=*]

  \item We propose a grey-box adversarial attack framework against on-device models that adopt transfer learning. It is a complete, practical and scalable pipeline including model extraction, pre-trained model identification and model fine-tuning with the consideration of the model-specific characteristics, followed by an adversarial attack. Such a mechanism enables attackers to construct tailored adversarial examples for real-world on-device models, which remain effective regardless of various model structures and transfer learning strategies.

  \item We conduct an extensive evaluation on its performance and generality in different settings.  We also carry out a study on real-world deep learning mobile apps crawled from Google Play and succeed in 38 of them. 

  \item We provide justification for the generality and scalability of such attacks and discuss potential countermeasures. This analysis suggests that improving the robustness of fine-tuned models is necessary for developing deep learning mobile apps, pointing to several promising research directions.

\end{itemize}

\section{Background}
\label{sec:background}

\subsection{On-device Deep Learning Model}
Currently, deep learning powered mobile apps can be roughly categorized into two ways: cloud-based inference and on-device inference.
The key difference between these two architectures is the storage location of the deep learning model.
In cloud-based deep learning models, mobile devices need to send requests to a cloud server and retrieve the inference results.
Offloading inference execution to the cloud comes with multiple drawbacks such as data privacy
concern, unreliable network conditions, and high latency.
In comparison, on-device deep learning models avoid the aforementioned drawbacks of the cloud-based approach.
It performs real-time inference on smartphones without network connection, and rarely requires sending user data off the device.

The implementation of on-device deep learning models is often with the aid of deep learning frameworks, such as Google TensorFlow (TF) and TFLite \cite{tensorflowlite}, Facebook PyTorch \cite{pytorch} and Pytorch Mobile \cite{pytorchmobile}, and Apple Core ML \cite{coreml}.
Among these frameworks, TFLite is the most popular technology used for running deep learning models on mobile, embedded, and IoT devices. It contributes nearly half of the deep learning mobile apps in the past two years and its usage is growing more significantly than other frameworks \cite{huang2021robustness,xu2019first}.
Although existing deep learning frameworks reduce the engineering efforts of implementing on-device deep learning models, training a new deep learning model from scratch is still expensive.
Hence, pre-trained deep learning models and transfer learning techniques are usually utilized in deep learning mobile apps to reduce training costs \cite{huang2021robustness,ccurukouglu2018deep}.
This allows mobile developers to leverage the representations learned by a pre-trained network and fine-tune them to a specific task.

\subsection{Transfer Learning}

Transfer learning is proposed to transfer the knowledge from a pre-trained model to a new model (i.e., fine-tuned model) that performs a different task. 
This knowledge normally refers to the model structure and parameters affiliated to the layers.
By leveraging existing knowledge, transfer learning enables researchers or industries to speed up the development of the new models even when their learning tasks are different from the pre-trained models.

\begin{figure}[hbt!]
\vspace{-1.2em}
\centering
\includegraphics[width=\linewidth,height=4.7cm]{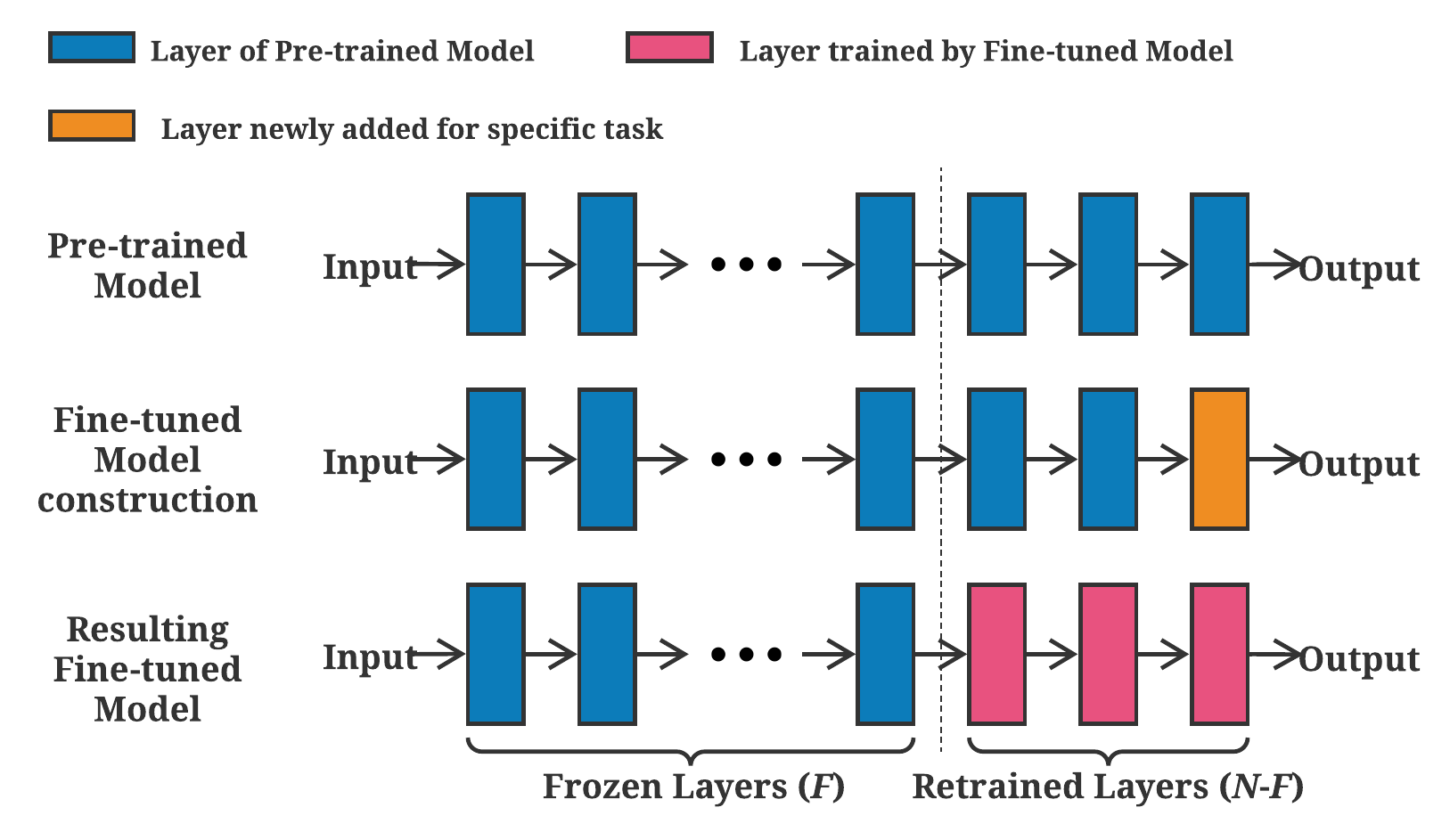}
\caption{Transfer learning.}
\label{fig:transfer_learning}
\vspace{-1.5em}
\end{figure}

\begin{figure*}[hbt!]
\vspace{-1.5em}
\centering
\includegraphics[width=\linewidth]{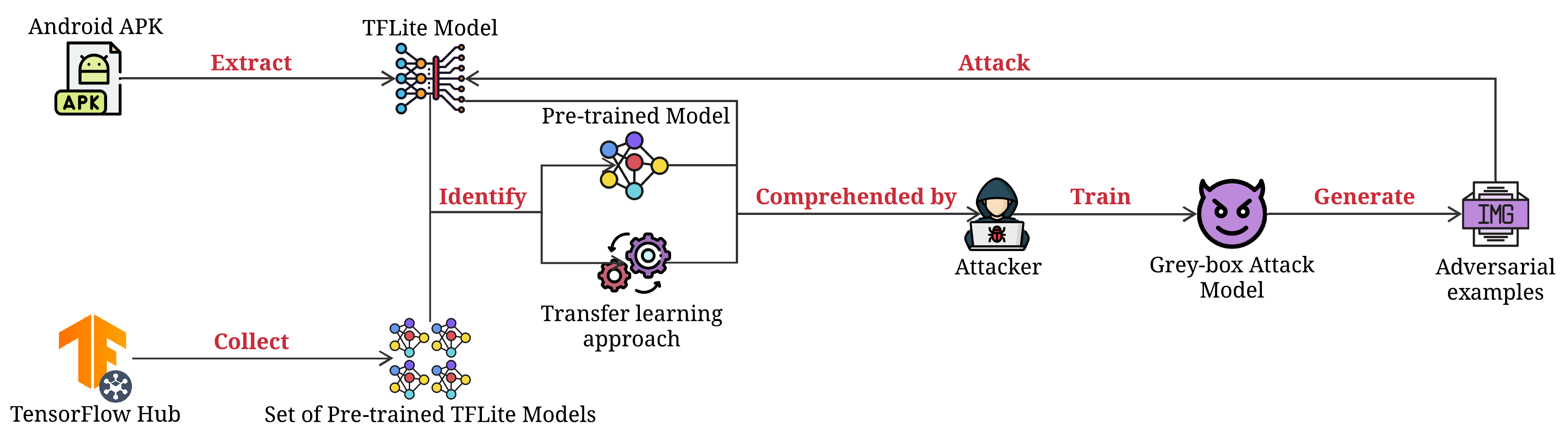}
\caption{The overall workflow of the attack framework.}
\label{fig:workflow}
\vspace{-1.5em}
\end{figure*}

Figure \ref{fig:transfer_learning} illustrates the process of transfer learning at a high level.
During initialization, a pre-trained model with $N$ layers is selected as the base model.
The fine-tuned model copies the first $N-1$ layers (i.e., both the structure and parameters) of the base model and add a new fully-connected layer that is tailored to fit the new task.
Subsequently, the fine-tuned model trains the last $N-F$ layers without updating the first $F$ layers (i.e., frozen layers) based on its own dataset.
This allows the fine-tuned model to exploit learned feature maps of the pre-trained model to extract meaningful features from new samples and thus lowers the training cost.
Depending on the number of frozen layers ($F$) during the training, transfer learning approaches can be categorized as follows:

\begin{itemize}[leftmargin=*]

  \item \textit{Feature Extraction.}
  The fine-tuned model freezes the first $F$ layers of pre-trained model, where $F=N-1$, and only updates the parameters of the newly-added classifier layer during the training.
  This approach is suitable when the task domains of both the fine-tuned and pre-trained models are similar, and the new training dataset is small.

  \item \textit{Fine-Tuning.}
  Instead of training only the last layers of the fine-tuned model, this approach trains the top-level layers of the pre-trained model alongside the training of the newly-added classifier.
  In this case, the frozen layers $F$ is less than $N-1$, which allows the parameters to be tuned from generic feature maps to features associated specifically with the new dataset.
  \textit{Fine-Tuning} often outperforms \textit{Feature Extraction} when the fine-tuned model’s objective is different than the original task of the pre-trained model, and the new training dataset is large.

\end{itemize}

\subsection{Adversarial Attacks in Deep Learning}
Adversarial attacks modify the original images with subtle perturbations, which can mislead the deep learning model to make incorrect predictions while hardly be distinguished by human eyes.
Based on the attacker's knowledge of the targeted classifier, existing attacks for crafting adversarial images fall into two categories.

\textbf{White-box Attacks.}
In the white-box setting, the attacker is assumed to fully know the targeted model, including both the structure and parameters \cite{goodfellow2014explaining}.
This type of attack~\cite{goodfellow2014explaining,carlini2017towards,rauber2020fast} allows the attacker to directly compute the perturbations that change the model prediction using gradient ascent and often achieve nearly 100\% attack success rate with minimal perturbations.
However, the inaccessibility of models in real-world applications makes the white-box attack impractical. 

\textbf{Black-box Attacks.}
In the black-box setting, the attacker does not know the targeted model (i.e., structure and parameters) and only can perform queries on it \cite{ilyas2018black}.
Some black-box attacks attempt to estimate decision boundaries of the targeted model using its outputs then construct a substitute model for generating adversarial images \cite{papernot2017practical,papernot2016transferability}.
Other attacks execute queries on the targeted model to evaluate the success of adversarial images and improve them accordingly \cite{guo2019simple,andriushchenko2020square}.
Compared to the white-box scenario, the black-box attack seems more realistic, however, it typically has the additional constraint on query budget, which lowers the attack effectiveness \cite{guo2019simple}.

\section{Attack to On-device Models}

\label{sec:methodology}
In this section, we present a grey-box adversarial attack framework specifically for hacking on-device models.
The framework first performs model extraction to a mobile app with a deep learning model and check if it is a fine-tuned model.
By identifying its pre-trained model and corresponding transfer learning approach, the attacker can build a binary classification model (i.e., adversarial model) against the targeted model (i.e., fine-tuned model) to craft adversarial images that fool the targeted model to misclassify a specifically targeted class.
The overall workflow of our attack framework is depicted in Figure \ref{fig:workflow} and the details of training Grey-box Attack Model are depicted in Figure \ref{fig:bin_training}.

\subsection{Extracting On-device Models}
\label{subsec:extract}
Since our attack framework is specifically for on-device TFLite models, to obtain targeted models from a mobile app (i.e., Android app), we first examine whether the mobile app adopts TFLite models, and if so, extract the models from it.
Given an Android APK, we utilize Apktool \cite{winsniewski2012apktool} to decompile it into nearly original form, including resource files, .dex files, manifest files, etc. 
After decompiling, we check whether the decomposed APK comprises files with the TFLite model naming schemes \cite{sun2021mind}, if any such exist, extract model files from it.
During extraction, each extracted model's completeness and operability are checked by loading the model and performing inference on randomly generated data.
This is vital for our attack as determining whether a targeted model is a fine-tuned model requires its full information (i.e., model structure and parameters).
In addition, the availability of targeted models is essential to perform the attack.

\subsection{Locating Fine-tuned Models}
\label{subsec:locate}
When locating a fine-tuned model, we consider two metrics: the structural similarity and the parametric similarity between a targeted model and a standard pre-trained model from TensorFlow Hub (TensorHub for short) \cite{TensorHub}.

\begin{figure}[hbt!]
\centering
\includegraphics[scale=0.5]{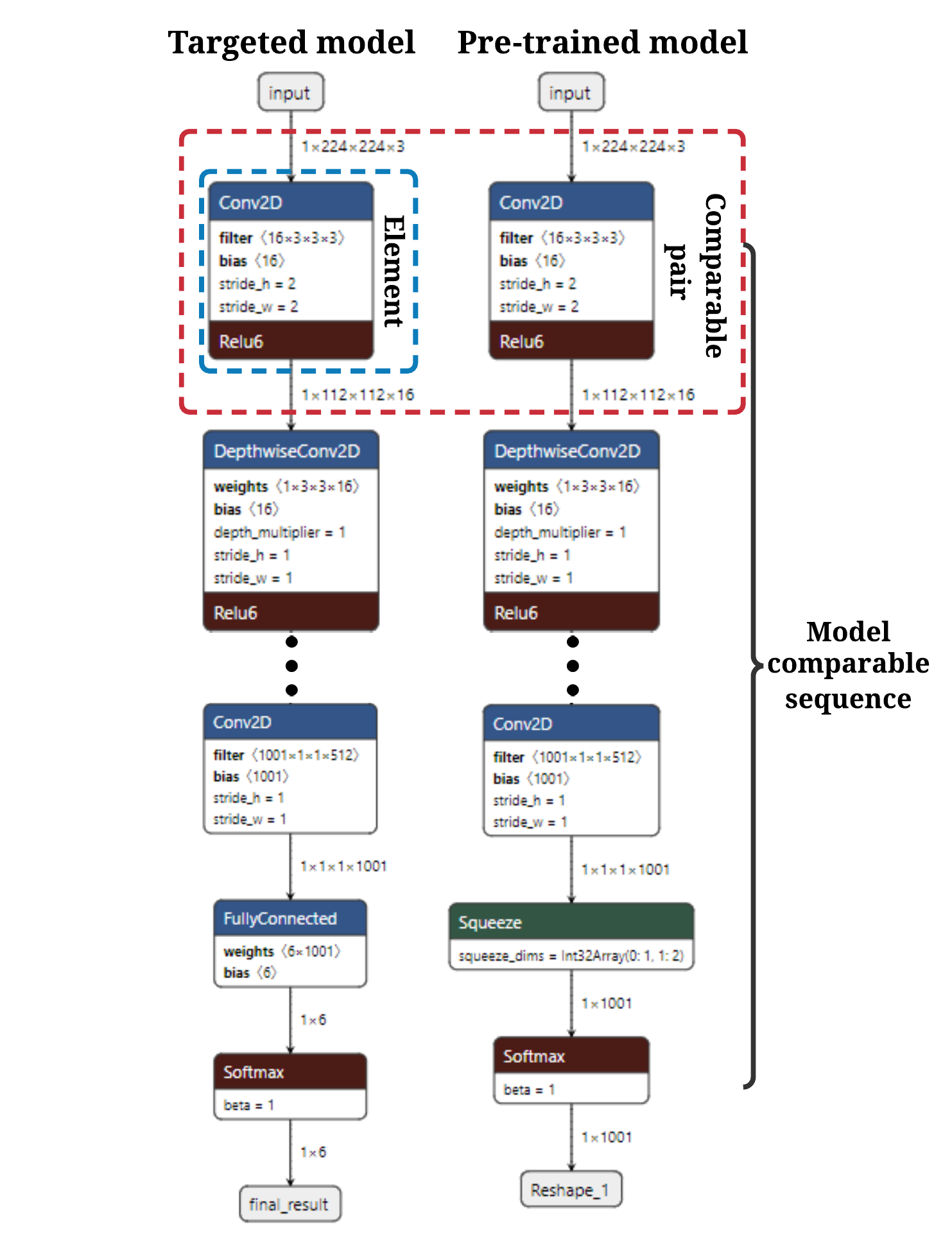}
\caption{Sequence conversion of TFLite model, visualized by Netron \cite{netron}.}
\label{fig:conversion}
\vspace{-1.5em}
\end{figure}

\textbf{Structural Similarity.}
Given a targeted model extracted from a deep learning mobile app, we first convert it to a sequence of elements for ease of comparison.
Each element within the sequence corresponds to one layer of the model and contains the layer's information, including identifier, shape, and data type.
As shown in Figure \ref{fig:conversion}, one convolutional layer of the targeted model is encoded to "MobilenetV1/Conv2d\_0/Relu6,[1,112,112,16],float32".
To obtain the structural similarity between a targeted model $M_{tar}$ and a pre-trained model $M_{pre}$ in TensorHub, we compute the Levenshtein distance \cite{levenshtein1966binary} between them.
The reason is that various targeted models may fine-tune pre-trained models by adding one or more new layers, resulting in model sequences of different lengths between targeted and pre-trained models, and the Levenshtein distance is able to compute the similarity for such sequences.
The calculation of structural similarity is formulated as follows:
\begin{equation}\label{eq:stru}
    stru\_sim(M_{tar}, \ M_{pre}) = \frac{L_{total} -LD(S_{tar},S_{pre})}{L_{total}}
\end{equation}
where $S_{tar}$ and $S_{pre}$ are respectively the model sequences of $M_{tar}$ and $M_{pre}$, $LD$ is the Levenshtein distance function, $L_{total}$ is the total number of layers of two models, and $stru\_sim$ is in the range of 0 to 1.
Note that an element is treated as a character (i.e., a sequence is treated as a string) in $LD$, and two elements in a comparable pair are considered as matched only if all attributes of them are identical.
Intuitively, higher structural similarity indicates that the two models are more similar in terms of structure.

For the structural similarity, the goal is to determine whether a targeted model takes advantage of the structure of a pre-trained model.
To find out an appropriate similarity threshold for determining that the two models (i.e., targeted and pre-trained models) are structurally similar, we conduct a pilot study on numerous real-world deep learning Android apps that adopt fine-tuned models to examine the structural similarities between these fine-tuned models and their corresponding pre-trained models.
Empirically, we find that 80\% structural similarity is a baseline for locating the pre-trained model in terms of structure.



\textbf{Parametric Similarity.}
In the context of our definitions in Section \ref{sec:background}, transfer learning is categorized into two approaches.
To ensure the effectiveness of our attack, we further adopt the parametric similarity to determine which transfer learning approach a fine-tuned model utilizes.
This information will be used to craft a binary adversarial model against the targeted model, as explained in Section \ref{sec:cra_adv_model}.
Given a fine-tuned model (i.e., targeted model) and its pre-trained model, $M_{fin}, M_{pre}$, with the structural similarity greater than or equal to 80\%, we still convert two models into sequences, with each layer as one element, but we adopt the detailed parameters of that layer as the attribute.
Subsequently, we perform subtractions on each comparable pair, i.e., two elements in the same position.
The calculation resulting in 0 represents two elements’ parameters are the same, then a Boolean value True will be stored in a sequence with the order.
Otherwise, a False will be stored.
The calculation of parametric similarity is formulated as follows:
\begin{equation}\label{eq:para}
    para\_sim(M_{fin}, \ M_{pre}) = \frac{N_{true}}{N_{total}}
\end{equation}
where $N_{true}$ is the number of longest continuous True values, $N_{total}$ is the total number of Boolean values in the resulting sequence, and $para\_sim$ is in the range of 0 to 1.
Based on the parametric similarity between $M_{fin}$ and $M_{pre}$, we can identify the transfer learning approach that the fine-tuned model adopts.
For instance, we follow the TensorFlow official transfer learning tutorial~\cite{tensorflowtra} to build a fine-tuned model based on a pre-trained MobileNetV2 that has 156 layers.
By freezing all layers of the base model (i.e., the pre-trained model without classification head has 154 layers) and only training the parameters of the newly-added classifier layers, the resulting fine-tuned model and its pre-trained model have a parametric similarity of approximately 98.72\%.

\subsection{Crafting Adversarial Models}
\label{sec:cra_adv_model}
In the proposed attack framework, attackers have white-box access to pre-trained models and partial white-box access to on-device models that adopt transfer learning (i.e., fine-tuned models) as they cannot be used to train adversarial examples. 
Hence, we consider a given attacker looking to trigger misclassifications of a particular class from a fine-tuned model $M_{fin}$, which has been customized through transfer learning from a pre-trained model $M_{pre}$.
The adversarial examples are crafted via a binary adversarial model $M_{bin}$ retrained from the $M_{pre}$.

\textbf{White-box Pre-trained Model.}
$M_{pre}$ is a white-box, meaning the attacker knows its model structure and parameters.
Most popular models, such as MobileNet \cite{howard2017mobilenets}, Inception \cite{szegedy2016rethinking}, and ResNet \cite{he2016deep}, have been made publicly available in various repositories (e.g., TensorHub) to increase adoption.
This allows attackers to obtain pre-trained models with minimal effort. 
Even if some pre-trained models are proprietary, attackers can still access them by pretending to be developers.

\textbf{Partial White-box Fine-tuned Model.}
The attacker is able to obtain the $M_{fin}$ and examine its structure and parameters via decompilation of the deep learning mobile app.
However, the training of adversarial examples cannot be directly performed on $M_{fin}$ due to the characteristics of the TFLite model. 
This causes the $M_{fin}$ being a partial white-box for the attacker.
Apart from knowing the characteristic of $M_{fin}$, the attacker can also access the corresponding label file stored in the decompiled app archive, which offers a chance to find the potential training datasets.

\textbf{Transfer Learning Parameters.}
We assume an on-device model adopts transfer learning.
Through the computation of structural and parametric similarities between it and each pre-trained model in TensorHub, the attacker is able to know the type of pre-trained model the on-device model adopted, and which layers were frozen during the fine-tuning, i.e., which transfer learning approach is adopted.
This information will be used in the construction of a binary adversarial model.

\textbf{Binary Adversarial Model.}
For simplicity, we assume the attacker attempts to fool an on-device model $M_{fin}$ that adopts transfer learning to misclassify a targeted image into any class other than the true one.
As the attacker is unable to perturb the targeted image on the $M_{fin}$ but has knowledge of its pre-trained model $M_{pre}$ and transfer learning approach $A_{tra}$, we thus consider crafting a binary adversarial model $M_{bin}$ that is highly similar to the $M_{fin}$ based on the $M_{pre}$ and $A_{tra}$.

To trigger the $M_{fin}$ to misclassify targeted images $x_+$ into any class $l_-$ different from the targeted class $l_+$, we first collect a set of inputs divided into two parts: targeted images $x_+$ and non-targeted images $x_-$ (i.e., the images correspond to one of the other classes recognized by $M_{fin}$ except for the targeted one).
We then feed the collected images (i.e, $x_+$ and $x_-$) to the $M_{fin}$ to select correctly classified inputs and divide them into the training and test sets ($X_{train}$ and $X_{test}$) for crafting the $M_{bin}$.
To obtain the optimal $M_{bin}$ for constructing adversarial images, we define the optimization problem below:

\begin{equation} \label{eq1}
\begin{split}
\max & \quad s(M_{bin},M_{fin}),\; s_{stru},s_{para} \in s \\
\min & \quad \ell(f,X_{train}) \\
\mathrm{s.t.} & \quad f(X_{test}) \geq f'(X_{test})
\end{split}
\end{equation}

where $s$ is the similarity function of model structure (formula \ref{eq:stru}) and parameters (formula \ref{eq:para}), $\ell$ is the sparse categorical cross entropy loss function,  and $f$ and $f'$ are respectively the classifier function of $M_{bin}$ and $M_{fin}$.
The above optimization attempts to maximize the structural and parametric similarities between $M_{bin}$ and $M_{fin}$ and minimize the loss of $f$ on $X_{train}$, under a constraint that the accuracy of $f$ on $X_{test}$ is at least the same as that of $f'$.
The key insight is that the attacker can compute perturbations for the $x_+$ based on the $M_{bin}$, and use the modified $x{_+}'$ (i.e., misclassified as $l_-$ by $M_{bin}$) to attack $l_+$ on the $M_{fin}$ as the two models are highly similar in terms of the targeted features.



\begin{figure}[hbt!]
\centering
\includegraphics[width=\linewidth]{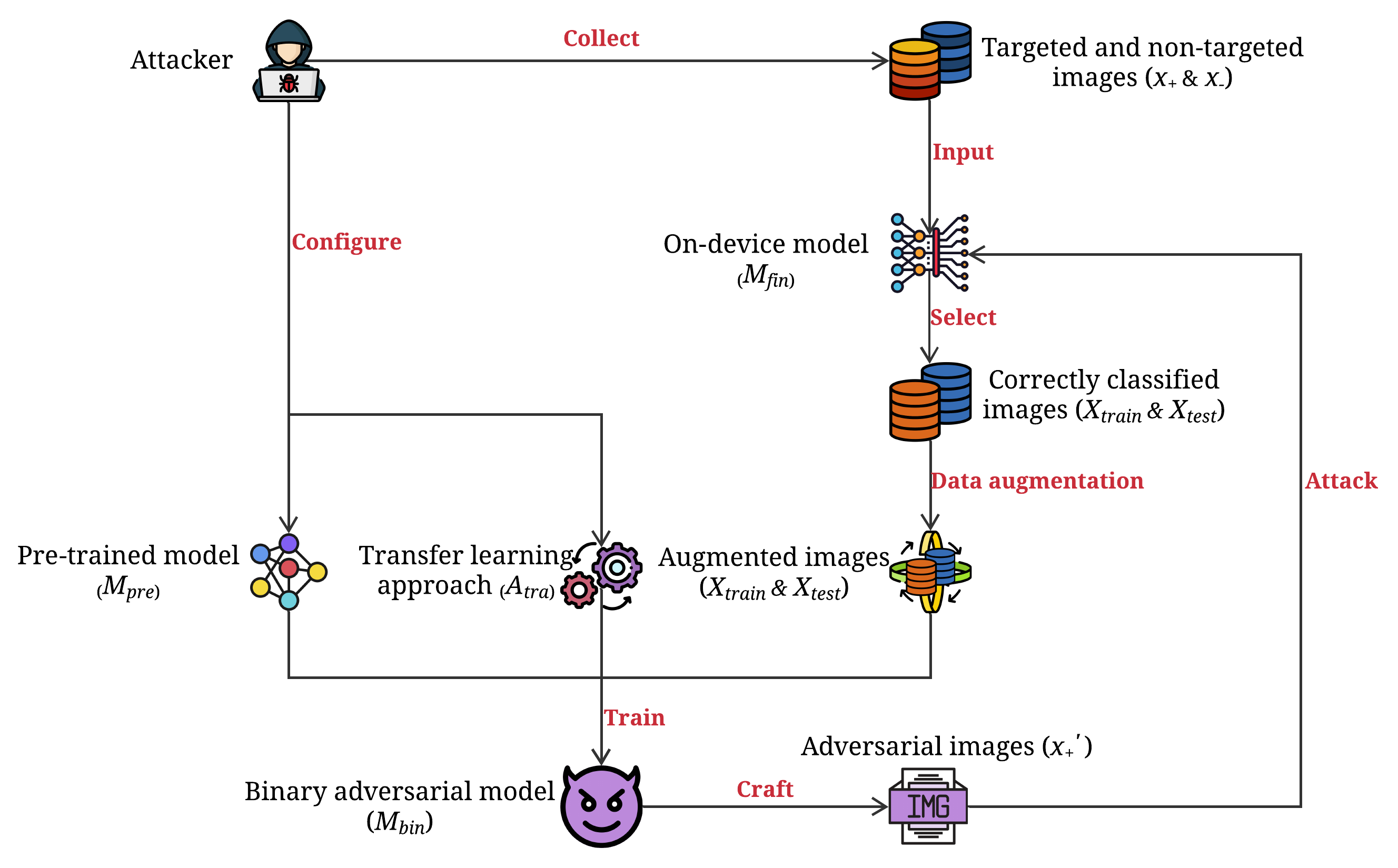}
\caption{Binary adversarial model training and corresponding adversarial images crafting.}
\label{fig:bin_training}
\vspace{-1.4em}
\end{figure}

One critical challenge encountered during the binary model training is that the training data is insufficient compared to the original data used for model fine-tuning, which may cause the $M_{bin}$ overfitting and reduce the power of generalization (i.e., the performance on the targeted class is not similar to the $M_{fin}$).
In this case, the adversarial images generated on $M_{bin}$ are less powerful to the $M_{fin}$.
To overcome that limitation, we adopt the data augmentation \cite{perez2017effectiveness} to artificially augmenting the existing dataset by applying random, yet realistic, transformations to the training images, such as rotation and horizontal flipping.
This helps expose the $M_{bin}$ to different aspects of the training data and thus enhance the ability of adversarial examples.
The complete process of training a binary adversarial model and crafting corresponding adversarial images is shown in Figure \ref{fig:bin_training}.

\vspace{-1em}

\begin{table}[hbt!]
\centering
\small
\setlength{\tabcolsep}{4pt}
\begin{tabularx}{\linewidth}{cccclcc} 
\noalign{\hrule height 1pt}
Label & $l_+$ (no passing) & $l_2$ & \boldsymbol{$l_3$} (\textbf{selected as} \boldsymbol{$l_e$}) & ... & $l_n$ \\ 
\hline
Classified count & 532 & 12 & \textbf{32} & ... & 8 \\
\noalign{\hrule height 1pt}
\end{tabularx}
\caption{Error matrix of the traffic sign recognition model with respect to the input "no passing".}
\label{table: error_matrix}
\vspace{-1em}
\end{table}

\begin{figure}[hbt!]
\centering
  \vspace{-2em}
  \subfloat[Targeted class image.]{
	\begin{minipage}[c][0.6\width]{
	   0.45\linewidth}
	   \centering
	   \includegraphics[width=0.5\linewidth]{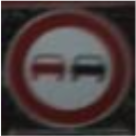}
	\end{minipage}}
 \hfill 	
  \subfloat[Most error-prone class image.]{
	\begin{minipage}[c][0.6\width]{
	   0.45\linewidth}
	   \centering
	   \includegraphics[width=0.5\linewidth]{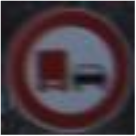}
	\end{minipage}}

\caption{Image examples from the targeted (no passing) and its most error-prone classes.}
\label{fig:traffic_error_prone}
\vspace{-1em}
\end{figure}

Besides the default setting, we also boost the attack success rate by finding the most error-prone class $l_e$ regarding $l_+$.
To this end, we introduce an error matrix (\boldsymbol{$E$}) to summarize the output of $M_{fin}$ when only passing $x_+$ as model input.
Then the objective is formulated as follows:
\begin{equation}
    l_e = \position_i (\max \boldsymbol{E}^{x_+}_{-j})
\end{equation}
where \boldsymbol{$E$}$^{x_+} \in \mathbb{R}^{1 \times n}$, $n$ is the total number of classes recognized by $M_{fin}$, $j$ is the index of $l_+$ in \boldsymbol{$E$}$^{x_+}$, and $i$ is the index of $l_e$ found in \boldsymbol{$E$}$^{x_+}$.
For instance, given a traffic sign recognition model and a targeted class "no passing", the resulting error matrix is shown in Table \ref{table: error_matrix}.
It is clear that $l_3$ is the most error-prone among the misclassified classes as its classified count (32) is much higher than other classes and thus selected as $l_e$.
Figure \ref{fig:traffic_error_prone} shows an example comprising images taken from the targeted class "no passing" and its most error-prone class.
Observe that the two classes are highly similar, with only the shape of the red car being slightly different.
Hence, by substituting the non-targeted images $x_-$ in the training set with the most error-prone images $x_e$, the newly generated adversarial images can be more effective for the $M_{fin}$.
The intuition is that the targeted class shares a part of special features with the most error-prone class, and thus pushing the targeted image outside its decision boundary (i.e., forcing $x_+$ into the decision boundary of $l_e$) becomes more effortless.

\section{Evaluation}
\label{sec:evaluation}
There are four settings that may influence the effectiveness of our attack framework against on-device models including different pre-trained models, datasets, transfer learning approaches and adversarial attack algorithms.
To demonstrate the attack success rate and generality, we carry out experiments in all four different settings by specifically targeting at one aspect while keeping the other three aspects the same.
For example, when evaluating the performance of our attack in terms of datasets, the selected transfer learning approach, pre-trained model, and adversarial attack are fixed.
Note that we adopt the variable control to avoid the explosion of setting combinations. 

\subsection{Experimental Setup}

Given one smart app, our approach first extracts the deep learning model.
It then identifies the pre-trained model and how the model fine-tunes the pre-trained model in different datasets. 
According to the targeted attack class, we then craft the binary model and generate the adversarial cases using different adversarial attacks.

\textbf{Pre-trained Models.}
In experiments, we use three different TensorFlow official pre-trained models including MobileNetV2 \cite{sandler2018mobilenetv2}, InceptionV3 \cite{szegedy2016rethinking} and ResNet50V2 \cite{he2016identity} to build our victim fine-tuned models (i.e., on-device models).
All the pre-trained models are trained on the ImageNet \cite{deng2009imagenet} dataset of 1.3 million images, these models can effectively serve as generic models of the visual world and are capable of transfer learning.

\textbf{Datasets.} 
Since most on-device models are commonly used in task domains related to the images \cite{xu2019first,huang2021robustness}, we follow the previous works \cite{wang2018great,abdelkader2020headless} to select three frequently-used image classification datasets to build the victim fine-tuned models for experiments. 
The classification tasks associated with these datasets represent typical scenarios developers may face during transfer learning.

\begin{itemize}[leftmargin=*]


  \item \textbf{Object Recognition.}
  It aims to classify an image of an arbitrary object into its corresponding class.
  The fine-tuned models are trained using the CIFAR-10 dataset \cite{krizhevsky2009learning} containing 50,000 object images from 10 classes, and comes with a test dataset of 10,000 images.
  
  \item \textbf{Traffic Sign Recognition.}
  Its objective is to classify various traffic signs based on images, which can be used by autonomous cars to automatically recognize traffic signs.
  The fine-tuned models are trained on the GTSRB dataset \cite{Stallkamp-IJCNN-2011} containing 39,209 images of 43 different traffic signs. It also includes a test dataset of 12,630 images.
  
  \item \textbf{Flower Recognition.}
  Its purpose is to classify images of flowers into different categories, which is a common example of multi-class classification.
  The fine-tuned models are trained on the Oxford Flowers dataset \cite{Nilsback08} containing 6,149 images from 102 classes, and the test dataset contains 1,020 images.
\end{itemize}

\textbf{Transfer Learning Approaches.}
To evaluate the effectiveness of our attack on two transfer learning approaches (discussed in Section 2), we unfreeze a different number of the top layers (except for the classifier) of a pre-trained model (e.g., MobileNetV2) and jointly train both the newly-added classifier as well as the last unfreezing layers of the base model to build our victim fine-tuned models.
These resulting models are able to cover most tuning strategies.

\textbf{Adversarial Attack Algorithms.}
For the evaluation of our attack effectiveness against different adversarial attacks, we focus on untargeted attacks in the white-box setting as our attack fools fine-tuned models to misclassify targeted images by constructing adversarial examples on known binary adversarial models. 
Considering a wide range of white-box untargeted attack algorithms have been proposed, it is unfeasible to cover all of them.
We thus select three representative attacks including Fast Gradient Sign Method (FGSM)~\cite{goodfellow2014explaining}, Carlini and Wagner (C\&W)~\cite{carlini2017towards}, and Clipping-Aware Noise (CAN)~\cite{rauber2020fast} attacks for experiments as they are either the basis of many powerful attacks or effective in computer vision tasks \cite{ye2020detection,rusak2020simple}.

\textbf{Baselines}
To attack the victim fine-tuned models, we adopt three different settings:
(1) Default Binary Adversarial Model Attack ($BAMA$), which crafts adversarial images based on a binary model trained on the targeted class (i.e., the class the attacker intends to force the victim model to misclassify) and non-targeted class (i.e., an arbitrary class recognized by the victim model except for the targeted one).
(2) Enhanced Binary Adversarial Model Attack ($E-BAMA$), it is similar to the first setting but substitutes the non-targeted class with the most error-prone class (i.e., the class most likely to be misclassified as the targeted one) during binary model training.
(3) Pre-trained Model Attack ($PMA$), which directly generates adversarial images solely based on the victim model's pre-trained model without taking any other model information into account, i.e., it ignores the structure and parameter information of a victim model.
Note that the last setting introduced by \cite{huang2021robustness} is used to perform adversarial attacks for fine-tuned on-device models, we thus consider comparing it with the proposed attacks (i.e., $BAMA$ and $E-BAMA$).

\textbf{Attack Configuration.}
We craft adversarial examples using correctly classified images collected from the Internet.
These images are not seen by the fine-tuned model during its training and align with our attack setting, i.e., the attacker has no access to the training dataset of the on-device model, and the newly collected images simulate user input.
In each set of experiments, we collect 50 source images for a targeted class to perform attacks. 
Success for the untargeted attack is measured as the percentage of source images that are incorrectly classified into any other arbitrary class, so
\begin{equation}
    ASR = \frac{m}{t}
\end{equation}
where $m$ represents the number of misclassified images, $t$ is equal to 50, and $ASR$ represents the attack succes rate.

To evaluate our attack from the above four perspectives, we control the variable i.e., checking attack performance by changing one variable while keeping the other three ones fixed. 
We implement the attack using TensorFlow Lite \cite{tensorflowlite}, leveraging the open-source implementation of adversarial attacks provided by the prior work \cite{rauber2017foolbox}.
In experiments, the fine-tuning uses the cross-entropy loss and the Adam optimizer with the default setting as: learning rate = $10^{-3}$, $\beta_{1}=0.9$, and $\beta_{2}=0.999$.
The results with respect to each perspective are presented in the following.

\subsection{Effectiveness of the Attack}
\label{sec:att_eff}
\textbf{Performance on different pre-trained models.}
We build three different victim fine-tuned models based on pre-trained MobileNetV2, InceptionV3 and ResNet50V2 by using the Oxford Flowers dataset and \textit{Feature Extraction}.
The reason for choosing this dataset is that transfer learning is mainly used as a solution to the data scarcity problem (i.e., small training dataset) \cite{wang2018great,zhang2020two}, and the Oxford Flowers dataset contains less data compared to the other two.
After fine-tuning, three victim models attain 86.08\%, 73.73\% and 82.45\% accuracy, respectively.
To construct adversarial images for a randomly selected targeted class (i.e., Bougainvillea in our experiments), we adopt the CAN attack with the epsilon = $20$ for $PMA$, $BAMA$ and $E-BAMA$.
These values are experimentally derived by authors to produce unnoticeable image perturbations.
For constructing $E-BAMA$, we use the error matrix mentioned in Section \ref{sec:cra_adv_model} to find out the most error-prone class Anthurium for Bougainvillea.

\begin{table}[hbt!]
\vspace{-0.5em}
\centering
\footnotesize
\resizebox{\linewidth}{!}{%
\begin{tabular}{c|ccc} 
\noalign{\hrule height 1pt}
\backslashbox[20mm]{Setting}{Model}  & MobileNetV2~ & InceptionV3 & ResNet50V2  \\ 
\hline
$PMA$       & 0.14       & 0.04      & 0.06      \\ 

$BAMA$       & 0.36       & 0.12      & 0.14      \\ 

$E-BAMA$ & \textbf{0.44}       &  \textbf{0.22}     &  \textbf{0.2}      \\
\noalign{\hrule height 1pt}
\end{tabular}}
\caption{Attack success rate on Flower Recognition against different pre-trained models.}
\label{table: pre_trained_impact}
\vspace{-1.1em}
\end{table}

Table \ref{table: pre_trained_impact} summarizes the impact of pre-trained model selection on the attack effectiveness.
Compared with the default setting ($BAMA$), the success rate of our attack is at least 133\% higher than that of $PMA$ for different types of fine-tuned models.
The poor performance of $PMA$ can be attributed to the significant difference between the fine-tuned and pre-trained models' tasks, thus adversarial images crafted using decision boundary analysis of the pre-trained model fail on the fine-tuned model.
Moreover, when adopting the enhanced setting ($E-BAMA$), the attack success rate is further improved by approximately 6-10\%.
In particular, we find that the fine-tuned model of MobileNetV2 is the most vulnerable compared to the other two, where the attack success rate of $BAMA$ and $E-BAMA$ are 36\% and 44\%, respectively.
This can be explained by the fact that when a fine-tuned model has higher accuracy, it is more sensitive to perturbations.
Finally, the results suggest that the attack effectiveness is not correlated with the pre-trained model: our attack framework is effective for various fine-tuned models irrespective of the types of their pre-trained models.

\textbf{Performance on different datasets.}
We pair the feature extractor of the pre-trained MobileNetV2 with a newly-added classifier head and adapt them to the CIFAR-10, GTSRB and Oxford Flowers datasets to build three victim fine-tuned models.
Such models achieve 81.79\%, 82.70\% and 86.08\% accuracy, respectively.
For each dataset (i.e., recognition task), we randomly pick a targeted class and locate its most error-prone class via the corresponding error matrix. 
The results are shown in Table \ref{table: targeted_error-prone}.
To craft adversarial images, we still use the CAN attack with the same setting (i.e., epsilon =20) as we only explore the attack effectiveness against different datasets.

\begin{table}[hbt!]
    \vspace{-0.5em}
	\centering
	\tiny
	\resizebox{\linewidth}{!}{%
		\begin{tabular}{c|ccc} 
			\noalign{\hrule height 0.7pt}
			\backslashbox[18mm]{Dataset}{Class}    & Targeted  & Most error-prone \\ 
			\hline
			CIFAR-10       & Airplane       & Truck                  \\ 
			
			GTSRB          & Stop           & No passing             \\ 
			
			Oxford Flowers & Bougainvillea & Anthurium              \\
			\noalign{\hrule height 0.7pt}
	\end{tabular}}
	\caption{The selected targeted class and its most error-prone classs for each dataset. }
	\label{table: targeted_error-prone}
	\vspace{-1.9em}
\end{table}

Table \ref{table: dataset_impact} shows how the attack effectiveness varies with respect to different datasets.
Across all the recognition tasks, both $BAMA$ and $E-BAMA$ outperform $PMA$, with more than 112\% higher attack success rate. 
Specifically, we find that the attack success rates of $BAMA$ and $E-BAMA$ are above 70\% for Traffic Sign Recognition.
This is because the classes inside the GTSRB dataset are relatively similar, then our binary adversarial model can more easily make the targeted class to be misclassified into an arbitrary class instead of the true one.
Based on these results, it is clear that our attack framework remains effective regardless of various datasets and can lead to higher misclassification errors if the targeted recognition task containing similar classes.

\begin{table}[hbt!]
\vspace{-0.5em}
\centering
\footnotesize
\resizebox{\linewidth}{!}{%
\begin{tabular}{c|ccc} 
\noalign{\hrule height 1pt}
\backslashbox[20mm]{Setting}{Dataset} & CIFAR-10 & GTSRB  & Oxford Flowers  \\ 
\hline
$PMA$       & 0.06   & 0.34 & 0.14                                              \\ 

$BAMA$       & 0.14   & 0.72 & 0.38                                              \\ 

$E-BAMA$ &  \textbf{0.24}   & \textbf{0.78} &  \textbf{0.48}                                              \\
\noalign{\hrule height 1pt}
\end{tabular}}
\caption{Attack success rate of MobileNetV2's fine-tuned models on various datasets.}
\label{table: dataset_impact}
\vspace{-1.2em}
\end{table}

\begin{figure*}[hbt!]
\vspace{-1.5em}
\centering
\includegraphics[scale=0.6]{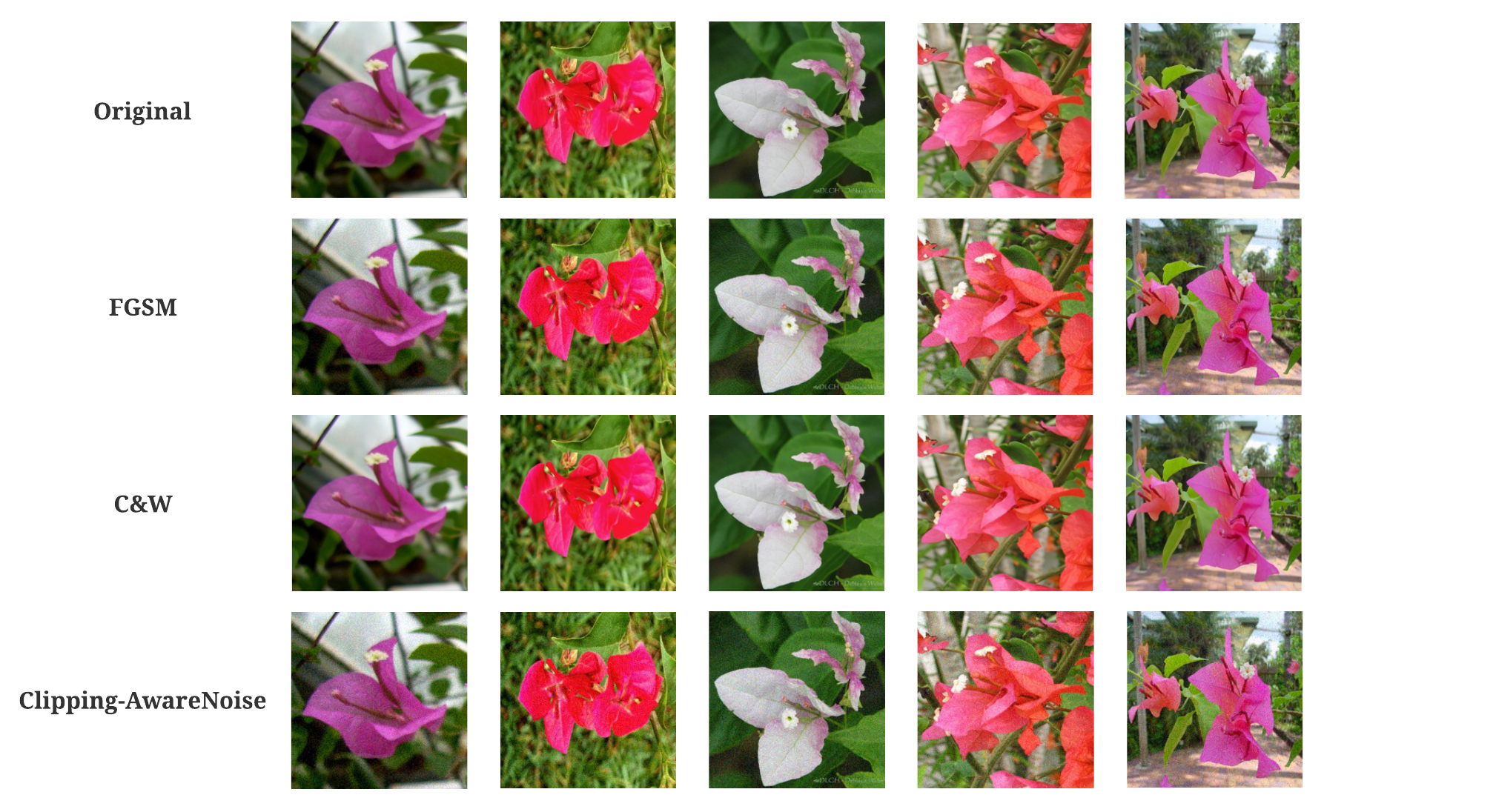}
\caption{Image of original Oxford Flowers and adversarial images generated by FGSM, C\&W and CAN attacks.}
\vspace{-1.5em}
\label{fig:adv_imgs}
\end{figure*}

\begin{figure}[hbt!]
\vspace{-2.5em}
\centering
  \subfloat[Target class image.]{
	\begin{minipage}[c][1\width]{
	   0.3\linewidth}
	   \centering
	   \includegraphics[width=0.8\linewidth]{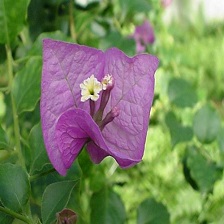}
	\end{minipage}}
 \hfill 	
  \subfloat[Human selection.]{
	\begin{minipage}[c][1\width]{
	   0.3\linewidth}
	   \centering
	   \includegraphics[width=0.8\linewidth]{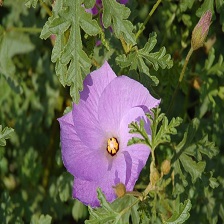}
	\end{minipage}}
 \hfill	
  \subfloat[Model selection.]{
	\begin{minipage}[c][1\width]{
	   0.3\linewidth}
	   \centering
	   \includegraphics[width=0.8\linewidth]{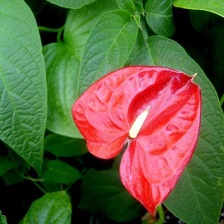}
	\end{minipage}}

\caption{Most error-prone class images selected by the human and model. The classes of (a), (b) and (c) are Bougainvillea, Pelargonium and Anthurium, respectively.}
\label{fig:error_prone}
\vspace{-1em}
\end{figure}

Apart from evaluating the attack, we find that the targeted and most error-prone classes are not very similar or even completely different.
This might be because of the visual perception difference between the human and model.
For instance, in the case of flower recognition, the human perceives whether two classes are similar based on the concrete object (e.g., the whole flower with color), while the model perception may be based on the characteristics of the object (e.g., the texture of the third flower is similar to that of the first), as shown in Figure \ref{fig:error_prone}.

\textbf{Performance on different transfer learning approaches.}
We keep the training dataset (Oxford Flowers), the victim fine-tuned models' architectures (i.e., MobileNetV2's base model plus a fully connected layer) and the attack setting (i.e., CAN attack with epsilon=20) the same in each set of trials.
But the number of unfreezing layers of the base model is set to a range of 0 to 60, with an interval of 10, based on the recommended setting of transfer learning provided by TensorFlow \cite{tentransfer}.
Note that 0 unfreezing layer means we adopt \textit{Feature Extraction} for transfer learning, while the other number of unfreezing layers represents different levels of \textit{Fine-Tuning}.

\begin{table}[hbt!]
\vspace{-0.5em}
\centering
\large
\resizebox{\linewidth}{!}{%
\begin{tabular}{c|c|cccccc} 
\noalign{\hrule height 1.8pt}
\multirow{2}*{\backslashbox[25mm]{Setting}{Layer}} & Feature Extraction & \multicolumn{6}{c}{Fine-Tuning}               \\ 

                  & 0                  & 10     & 20     & 30   & 40   & 50   & 60     \\ 
\hline
$PMA$              & 0.22      & 0.16   & 0.12   & 0.12 & 0.20  & 0.14 & 0.16  \\ 

$BAMA$             & 0.36  & 0.52 & 0.54  & 0.50  & 0.58 & 0.50  & 0.56   \\ 

$E-BAMA$           &  \textbf{0.48}  &  \textbf{0.58}   &  \textbf{0.62} &  \textbf{0.56} & \textbf{0.66} &  \textbf{0.60} &  \textbf{0.62}  \\
\noalign{\hrule height 1.8pt}
\end{tabular}}
\caption{Attack success rate of MobileNetV2's fine-tuned models against different number of fine-tuning layers.}
\label{table: layer_impact}
\vspace{-0.7em}
\end{table}

Table \ref{table: layer_impact} summarizes the attack effectiveness versus two transfer learning approaches.
Observe that as the number of fine-tuning layers varies (i.e., the transfer learning approach or level is changed), our attacks (either $BAMA$ or $E-BAMA$) remain higher attack success rates than $PMA$ regardless of \textit{Feature Extraction} or \textit{Fine-Tuning}, which indicates that the proposed attack framework is universal and robust against common cases of transfer learning.


\textbf{Performance on different adversarial attacks.}
We consider three types of adversarial attack algorithms, FGSM, C\&W and CAN.
Their epsilons are set to 0.025, 0.2 and 20, respectively.
These selected epsilon values are experimentally derived by authors,
so as to make the generated adversarial images preserve high credibility as far as possible.
Apart from epsilons, the rest of the experimental settings (i.e., the pre-trained model MobileNetV2, the Oxford Flowers dataset and \textit{Feature Extraction}) remain the same.

\begin{table}[hbt!]
\vspace{-0.5em}
\centering
\setlength{\tabcolsep}{13pt}
\footnotesize
\resizebox{\linewidth}{!}{%
\begin{tabular}{c|ccc} 
\noalign{\hrule height 1pt}
\backslashbox[25mm]{Setting}{Attack} & FGSM & C\&W   & CAN  \\ 
\hline
$PMA$    & 0.16 & 0.04 & 0.14                                    \\ 

$BAMA$   & 0.34  & 0.12 & 0.36                                    \\ 

$E-BAMA$ &  \textbf{0.40} &  \textbf{0.20} & \textbf{0.46}                                     \\
\noalign{\hrule height 1pt}
\end{tabular}
}
\caption{Attack success rate of MobileNetV2's fine-tuned models under different adversarial attack algorithms.}
\label{table: attack_impact}
\vspace{-1.4em}
\end{table}

Table \ref{table: attack_impact} summarizes how the setting of attack algorithm selection influences the attack effectiveness.
For the FGSM and CAN attacks, $BAMA$ and $E-BAMA$ yield similar attack performance, with above 2 times attack success rates compared to $PMA$.
However, when adopting the C\&W attack, the attack success rates of all three settings (i.e., $BAMA$, $E-BAMA$ and $PMA$) are relatively low.
To understand the reason for such results, we examine the adversarial images generated by three attack algorithms and compare them with the original images, as shown in Figure \ref{fig:adv_imgs}.
Observe that the images created by the C\&W attack look nearly the same as the original inputs, only with a slight blur around the flowers, while the images crafted using FGSM and CAN attacks contain tiny perturbations if examining carefully.
These imperceptible perturbations may be responsible for the higher attack success rate.

\begin{table*}[hbt!]
\centering
\vspace{-1.5em}
\setlength{\tabcolsep}{8pt}
\tiny
\resizebox{\linewidth}{!}{%
\begin{tabular}{lccc} 
\noalign{\hrule height 0.7pt} \multicolumn{1}{c}{\textbf{Task domain}} & \textbf{Pre-trained model} & \textbf{\# DL apps adopting transfer learning} & \textbf{\# Fine-tuned model} \\ 
\hline
\multirow{3}{*}{Image recognition} & MobileNetV1 and V2 & 23 & 24 \\

 & InceptionV3 & 2 & 2 \\

 & SqueezeNet & 1 & 1 \\ 
\hline
\multirow{2}{*}{Object detection} & COCO SSD MobileNet v1 & 9 & 9 \\

 & Google Mobile Object Localizer & 2 & 2 \\ 
\hline
Image segmentation & DeepLabv3 & 15 & 17 \\ 
\hline
Pose estimation & PoseNet & 1 & 1 \\ 
\hline
Total &  & 53 & 56\\
\noalign{\hrule height 0.7pt}
\end{tabular}}
\caption{Number of DL apps adopting transfer learning and corresponding fine-tuned TFLite models.}
\vspace{-3em}
\label{table:vlu_apps}
\end{table*}

Apart from the above settings, we also evaluate the performance of our attack in terms of different combinations of the four settings, i.e., pre-trained model, dataset, transfer learning approach and adversarial attack algorithm are all taken into account simultaneously.
The combinations used in the experiments are carefully selected to ensure diversity.
The experiment results show that our attack remains effective and robust regardless of different combinations of the four settings.
Detailed experiment settings and results are available at \href{https://github.com/Jinxhy/SmartAppAttack}{https://github.com/Jinxhy/SmartAppAttack}.

\section{Attacking Real-world Deep Learning Mobile Apps}
\label{sec:attack_app}

To estimate the potential damage of the attack framework to real-world deep learning mobile apps (DL apps for short), we implement a semiautomatic attack pipeline based on the methods introduced in Section \ref{sec:methodology} and conduct an empirical study to evaluate the proposed attack on a large set of Android apps collected from the official Google Play market.

\textbf{Collecting DL Apps.}
To find the targeted DL apps, we first crawl 63,112 mobile apps from Google Play.
These apps cover a variety of categories (e.g., Medicine, Auto \& Vehicles, and Finance) related to the image domain.
Subsequently, we filter the apps and extract the corresponding on-device models by adopting our attack pipeline mentioned in Section~\ref{subsec:extract}.
We obtain 394 apps with 1,345 TFLite models.
Note that most DL apps contain multiple models as they serve more than one deep-learning-based
feature, or require multiple models to work together to perform specific tasks.

\textbf{Identifying and attacking fine-tuned models in DL Apps.}
Following the approach mentioned in Section~\ref{subsec:locate}, we locate DL apps that fine-tune the public pre-trained models from TensorHub for their own purpose.
However, among these extracted TFLite models, we find that 1,120 of them are default Google Mobile Vision framework \cite{googlemobilevision} without any change, which concentrates on face, barcode, and text detection.
Hence, we exclude 280 DL apps that only use the Google Mobile Vision framework and pass the remaining 114 apps to the attack pipeline.
46.49\% (53/114) apps are of fine-tuned models, which covers various task domains such as object detection, image segmentation, and pose estimation with details seen in Table \ref{table:vlu_apps}.
Among these DL apps, nearly half of them perform tasks related to image recognition, and MobileNetV1 \cite{howard2017mobilenets} and V2 \cite{sandler2018mobilenetv2} are the most commonly used pre-trained models (24/56).

By knowing pre-trained models and label files of DL apps that adopt fine-tune models, our attack framework could successfully attack 71.7\% (38/53) of them.
Here success means that the generated adversarial images against given targeted classes can deceive the apps without being perceived by humans.
For the cases of failure, the reasons include (i) the label file is missing so we cannot select a targeted class to perform the attack; (ii) the dataset for training a binary adversarial model against rare tasks (e.g., prostate cancer recognition) is insufficient, affecting the attack performance.
Given the fact that the number of DL apps is growing rapidly \cite{xu2019first}, we believe the problem should raise concerns in both the industry and research community.


\textbf{Real-world Examples.}
Next, we discuss several real-world DL apps in more details to illustrate how to attack them with the proposed framework and present the corresponding consequences. 
We select three DL apps used in security-critical tasks, including skin cancer recognition, traffic sign recognition, and cash recognition.


\begin{itemize}[leftmargin=*]

  \item \textbf{Skin Cancer Recognition App.}
  Skin cancer recognition can classify the skin cancer types and help in early detection \cite{nasr2016melanoma}, which are vital to ensure a high survival rate in patients. In this app, a CNN model takes the user's skin image captured by a camera as the input and outputs whether he/she has skin cancer with corresponding suggestions (i.e., none for health skin or dermatologist visit for skin cancer).
  
  To launch the attack on the CNN model, we first use the attack pipeline to identify that MobileNetV1 \cite{howard2017mobilenets} is used as the pre-trained model. By computing the parametric similarity between the CNN model and its pre-trained model, we know that a fully unfrozen base model (i.e., a pre-trained model without classification head) is adopted for  \textit{Fine-Tuning}.
  Based on the label file stored in the APK, we select a targeted class (Melanoma) and collect the corresponding data using Google's Dataset Search\footnote{https://datasetsearch.research.google.com/}.
  During the data collection, we obtain the most error-prone class (Nevus) to ensure the optimal attack performance based on the error matrix, which is produced by feeding the targeted images to the CNN model.
  Given such information, we build a binary adversarial model to generate adversarial images.
  The subsequent misclassification attack achieves a success rate of 94\% (i.e., 47 out of 50 are misclassified), 56.67\% higher than the previous attack \cite{huang2021robustness}.

    \begin{figure}[hbt!]
    \vspace{-2em}
    \centering
    \subfloat[Normal input  with correct output]{
	\begin{minipage}[c][1.5\width]{
	   0.45\linewidth}
	   \centering
	   \includegraphics[width=\linewidth]{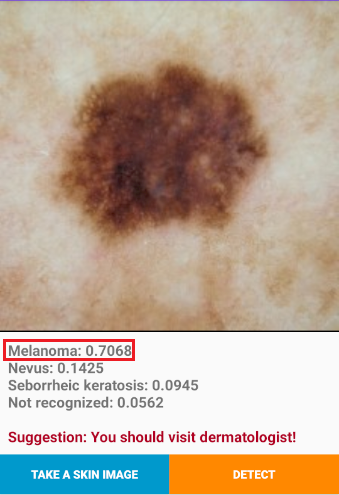}
	   \label{fig:skin_cancer_adv_normal_a}
	\end{minipage}}
    \hfill 	
    \subfloat[Adversarial input with wrong output]{
	\begin{minipage}[c][1.5\width]{
	   0.45\linewidth}
	   \centering
	   \includegraphics[width=\linewidth]{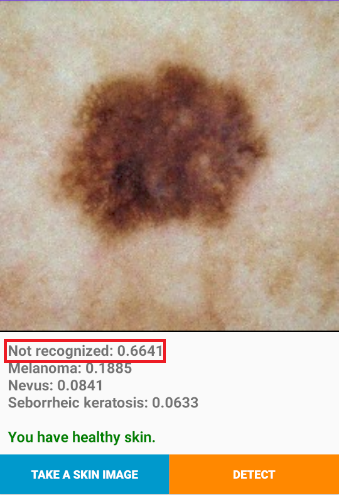}
	    \label{fig:skin_cancer_adv_normal_b}
	\end{minipage}}

    \caption{The behavior of skin cancer recognition app on normal and adversarial inputs.}
    \label{fig:skin_cancer_adv_normal}
    \vspace{-1.5em}
  \end{figure}
  

  Figure \ref{fig:skin_cancer_adv_normal_a} shows the app correctly classifies the skin lesion image as melanoma and suggests the user to see a dermatologist. 
  However, the second image generated by the binary adversarial model is recognized as healthy skin with a high confidence level.
  Due to the increasing use of deep learning to aid in digital health \cite{tan2019intelligent,nahata2020deep}, attacking such apps would affect medical professionals' judgement or directly pose threats to the end-users' lives.

  \item \textbf{Traffic Sign Recognition App.}
  Traffic sign recognition is able to detect and classify traffic signs for assisting driving.
  In this app, a sequence of video frames (images) captured by a camera are passed to a CNN model as the inputs.
  Once an important traffic sign (e.g., stop sign, no entry sign, and speed limit sign) is detected and recognized, the app would remind the driver to take a corresponding action (e.g., stopping, turning around, and reducing speed). 
  
  Again, we are able to launch the attack on the targeted CNN model via the attack pipeline.
  It identifies the pre-trained model used is MobileNetV1 \cite{howard2017mobilenets} and \textit{Feature Extraction} is adopted during transfer learning.
  In the subsequent training of the binary adversarial model, the collected data that contain the targeted class (Stop sign) and its most error-prone class (Do not overtake sign) are passed as inputs.
  Consequently, the misclassification attack achieves a 76\% (i.e., 38 out of 50 are misclassified) success rate, which is 2.3 times better than the previous attack \cite{huang2021robustness}.
  
  Figure \ref{fig:traffic_sign_adv_normal} shows the app behavior on normal and adversarial inputs.
  The app incorrectly reports the stop sign generated by our binary adversarial model as a speed limit sign.
  Since the self-driving vehicle technology is mainly backed up by deep learning \cite{grigorescu2020survey}, such apps may be directly used for controlling the vehicle in the future, attacking the models inside these apps would cause fatal incidents for end-users.

    \begin{figure}[hbt!]
    \vspace{-1.5em}
    \centering
    \subfloat[Normal input with correct output]{
	\begin{minipage}[c][1.1\width]{
	   0.45\linewidth}
	   \centering
	   \includegraphics[width=\linewidth]{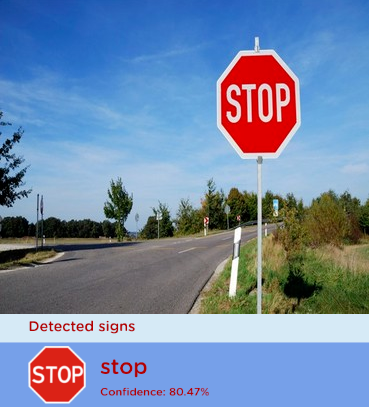}
	   \label{fig:traffic_sign_adv_normal_a}
	\end{minipage}}
    \hfill 	
    \subfloat[Adversarial input with wrong output]{
	\begin{minipage}[c][1.1\width]{
	   0.45\linewidth}
	   \centering
	   \includegraphics[width=\linewidth]{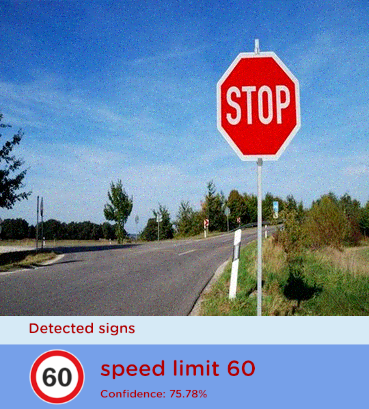}
	   \label{fig:traffic_sign_adv_normal_b}
	\end{minipage}}

    \caption{The behavior of traffic sign recognition app on normal and adversarial inputs.}
    \label{fig:traffic_sign_adv_normal}
    \vspace{-1em}
  \end{figure}

  \item \textbf{Cash Recognition App.}
  Cash recognition can classify various denominations of the currency (e.g., 20, 500, and 2000 Indian Rupee), which could be used to help visually impaired people to distinguish different monetary values.
  In this app, a user takes a picture of the cash, and the app reads the currency type and value to the user by using a customized model.
  
  Again we can launch the attack on the cash recognition model based on the information identified by our attack pipeline. 
  Here, the targeted model is built by retraining the top 28 layers of a MobileNetV1 \cite{howard2017mobilenets} base model along with the newly-added classification head (i.e., adopting \textit{Fine-Tuning}).
  We then select the targeted (500 Indian Rupee) and its most-error prone (100 Indian Rupee) classes to build a binary adversarial model to craft adversarial inputs.
  As a result, the misclassification attack achieves a success rate of 56\% (i.e., 28 out of 50 are misclassified), 143.48\% higher than the previous attack \cite{huang2021robustness}.
  
      \begin{figure}[hbt!]
    \vspace{-2em}
    \centering
    \subfloat[Normal input with correct output]{
	\begin{minipage}[c][0.8\width]{
	   0.45\linewidth}
	   \centering
	   \includegraphics[width=\linewidth]{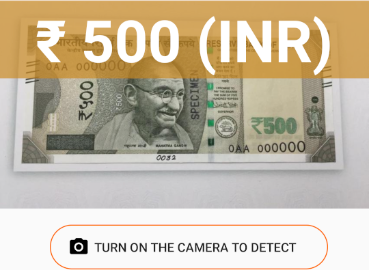}
	\end{minipage}}
    \hfill 	
    \subfloat[Adversarial input with wrong output]{
	\begin{minipage}[c][0.8\width]{
	   0.45\linewidth}
	   \centering
	   \includegraphics[width=\linewidth]{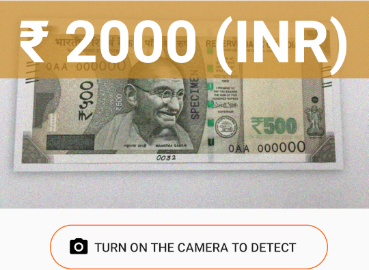}
	\end{minipage}}

    \caption{The behavior of cash recognition app on normal and adversarial inputs. \rupee{} and INR is the symbol and abbreviation of the Indian Rupee, respectively.}
    \label{fig:cash_adv_normal}
    \vspace{-0.6em}
  \end{figure}
  
  Figure \ref{fig:cash_adv_normal} shows the impact of feeding adversarial input to the app, in which an Indian 500-rupee banknote is misclassified as an Indian 2000-rupee banknote.
  Since visual impairment affects information acquisition, it is not hard to imagine that similar apps can be used in other types of accessibility services, such as reading newspapers, recognizing traffic conditions, etc.
  Attacking deep learning models of such apps could be a threat to people with visual impairment that relies on these accessibility services.

\end{itemize}

\section{Discussion}
\label{sec:discussion}
\subsection{Attack Generality}
In this paper, we demonstrate the effectiveness of our attack for on-device deep learning models related to the field of computer vision.
It is also possible to apply our attack to other fields, such as natural language processing and speech recognition.
For instance, given a sentiment analysis model based on a pre-trained neural-net language model, the attacker can first use our framework to identify its pre-trained model and train a binary adversarial model based on the targeted (e.g., positive sentiment) and non-targeted classes, the subsequent generated adversarial texts can be used to fool the targeted model (i.e., misclassify the positive text as negative or other sentiments).
Similarly, in speech recognition, the attacker can also adopt our framework to launch such attacks as long as the targeted models are based on transfer learning.

\begin{figure}[hbt!]
    \vspace{-1em}
    \centering
    \subfloat[Normal input with correct output]{
	\begin{minipage}[c][1.4\width]{
	   0.45\linewidth}
	   \centering
	   \includegraphics[width=\linewidth]{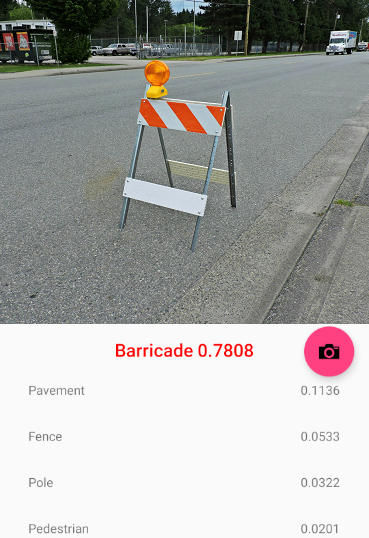}
	   \label{fig:obstacle_adv_normal_a}
	\end{minipage}}
    \hfill 	
    \subfloat[Adversarial input with wrong output]{
	\begin{minipage}[c][1.4\width]{
	   0.45\linewidth}
	   \centering
	   \includegraphics[width=\linewidth]{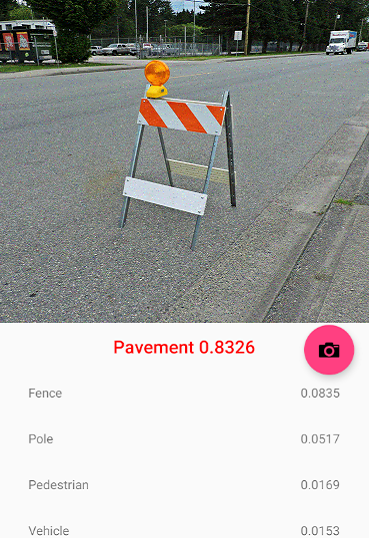}
	   \label{fig:obstacle_adv_normal_b}
	\end{minipage}}

    \caption{The behavior of obstacle recognition app on normal and adversarial inputs.}
    \label{fig:obstacle_adv_normal}
    \vspace{-1.5em}
\end{figure}

\begin{figure}[hbt!]
    \centering

    \subfloat[Normal input with correct output]{
	\begin{minipage}[c][1.2\width]{
	   0.45\linewidth}
	   \centering
	   \includegraphics[width=\linewidth]{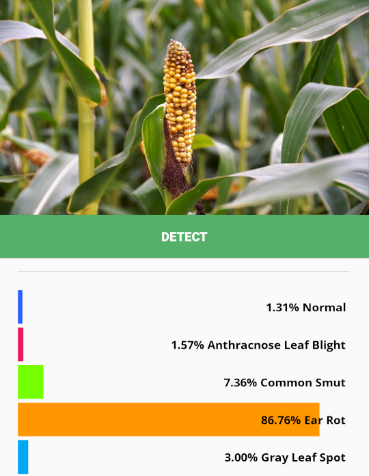}
	   \label{fig:crop_disease_adv_normal_a}
	\end{minipage}}
    \hfill 	
    \subfloat[Adversarial input with wrong output]{
	\begin{minipage}[c][1.2\width]{
	   0.45\linewidth}
	   \centering
	   \includegraphics[width=\linewidth]{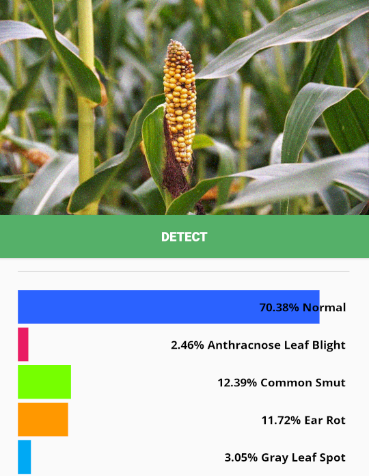}
	   \label{fig:crop_disease_adv_normal_b}
	\end{minipage}}

    \caption{The behavior of crop disease recognition app on normal and adversarial inputs.}
    \label{fig:crop_disease_adv_normal}
    \vspace{-1.8em}
\end{figure}

Besides the field generality, our proposed framework can also be easily adapted for attacking other edge-support machine learning frameworks such as PyTorch Mobile~\cite{pytorchmobile}.
To demonstrate the adaptability of the proposed framework, we replace TFLite-related model naming schemes (e.g., ".tflite"), pre-trained model repository (e.g., TensorHub) and framework APIs (e.g., "tf.lite.Interpreter") in the original attack framework with those (e.g., ".pt", Pytorch Hub\footnote{https://pytorch.org/hub/} and "torch.jit") related to PyTorch Mobile~\cite{pytorchmobile}.
We then apply our attack to two randomly-selected real-world DL apps that adopt PyTorch Mobile~\cite{pytorchmobile}.:
(1) obstacle recognition app that assists visually impaired people in distinguishing various obstacles in the form of reading the uploaded picture. It uses a MobileNetV2~\cite{sandler2018mobilenetv2} as the pre-trained model and adopts Feature Extraction for retraining.
(2) crop disease recognition app that helps farmers to recognize the disease of agricultural plants.
It adopts Fine-Tuning that retrains the top 20 layers of an InceptionV3~\cite{szegedy2016rethinking} base model along with the newly-added classification head.
The results are shown in Figure~\ref{fig:obstacle_adv_normal} and ~\ref{fig:crop_disease_adv_normal}.

\subsection{Attack Scalability}
On-device deep learning is not limited to mobile devices, it can also be applied to embedded and IoT devices, such as Raspberry Pi and microcontrollers \cite{vadlamudidriver,warden2019tinyml}.
Attacks on models inside such devices might be slightly different from those inside mobile devices, depending on the type of device.
For example, an embedded Linux device can directly run the locally stored model without installing APK~\cite{tfliteembedded}. 
In this case, the attacker can effortlessly obtain the model then adopt our framework to launch the attack.
If the model is deployed on a microcontroller, the attacker can recover it from the C byte array (i.e., the C source file that contains the model)~\cite{tflitemicrocontrollers} to the TFLite model and perform the attack.
Apart from this, it is also possible to extend our framework for attacking other edge devices that adopt on-device deep learning.

\subsection{Defence of Attack}
\label{discussion:defence}
To defend our attack, we discuss a few possible countermeasures from the perspective of practitioners.
First, deep learning mobile app developers are responsible for building
the on-device models, so they can take immediate and effective actions to secure the models, for instance: (1) encrypt the label file to prohibit attackers from collecting training data to build binary adversarial models.
(2) obfuscate the model information like layer names to prevent attackers from gaining insights into the model i.e., determining whether it is based on the pre-trained model. 
(3) adopt Android packing techniques to thwart attackers from obtaining any information associated with the model, such as the corresponding model framework API, model suffix convention and label file, thereby preventing them from building binary adversarial models.
(4) store and execute the model in secure hardware.
For instance, TF-Trusted~\cite{tftrusted} allows developers to run TensorFlow models inside of an Intel SGX device.
(5) train multiple fine-tuned models based on different types of pre-trained models for a given task, and use them together to make a prediction (i.e., majority vote). 
Thus even if the attacker successfully fools a single fine-tuned model in the ensemble, the other models may be resistant.
This is because our attack performance cannot remain the same when the types of pre-trained models changed, as discussed in Section \ref{sec:att_eff}.
For deep learning framework providers, they should consider providing better protection mechanisms: (1) signature-based model loading to ensure the model is utilized by trusted users. 
(2) built-in model encryption API assists developers to encrypt the models, avoiding attackers getting the model structures and parameters.

\section{Related Work}
\label{sec:related_work}
As this work focuses on the attack to the fine-tuned models in mobile apps, we introduce related works about the adversarial attacks to transfer learning and security of deep learning mobile apps in the following text.

\subsection{Adversarial Attacks in Transfer Learning}

Transfer learning has been demonstrated its effective to facilitate the effortless derivation of new models (i.e., fine-tuned models) through retraining a pre-trained model on a relevant task with less data and computation cost \cite{niu2021decade}.
However, pre-tained models are usually publicly available for sharing and reuse, which inevitably introduces vulnerability to trigger severe attacks (e.g., adversarial attacks) against transfer learning applications.

In recent years, researchers have proposed several novel adversarial attacks against deep neural networks in the context of transfer learning. 
For instance, Wang et al. \cite{wang2018great} proposed a transfer-based adversarial attack against Machine Learning as a service platforms, where the attacker is assumed to have knowledge about the pre-trained model and an instance of targeted image.
They exploit this knowledge to modify the internal representation of the source image to make it similar to the internal representation of the targeted image, which results in misclassification.
Similarly, Ji et al. \cite{ji2018model} crafted the malicious primitive model based on the pre-trained feature extractor, which is utilized to perturb the source image based on the targeted image such that their internal representations are close to each other. They exploit the transferability of adversarial examples to successfully attack corresponding fine-tuned models incorporated in machine learning systems.
Nevertheless, these attacks are only effective when victim fine-tuned models adopt \textit{Feature Extraction}, i.e., without changing any parameter in the pre-trained model's base model.
However, according to our observation, development teams may adopt different transfer learning approaches to fine-tune their models in mobile apps which make these attack invalid.


In addition, Prabhu and Whaley \cite{prabhu2018grey} utilized a pre-trained convolutional neural network model to generate adversarial examples, which successfully fool a black-box SVM classifier (i.e., it is not accessible to the attacker) that uses this pre-trained model as the feature extractor. 
Following this work, Rezaei and Liu \cite{rezaei2019target} demonstrated that without any additional knowledge other than the pre-trained model, attackers can effectively mislead a feature-extractor-based fine-tuned model by exploiting the vulnerabilities of Softmax layer.
Moreover, Abdelkader et al. \cite{abdelkader2020headless} found that using a known feature extractor (i.e., pre-trained model) exposes a fine-tuned model to powerful attacks that can be executed without knowledge of the classifier head at all. 
Recently, Pal and Tople \cite{pal2020transfer} exploited unintended features learnt in the pre-trained model to generate adversarial examples for fine-tuned models, which achieves a high attack success rate in text prediction task domain.

All these existing attacks assume pre-trained models are known to the attacker, which may not apply in real-world deep learning applications.
Our attack framework considers a much more realistic setting in which the attacker does not know the type of pre-trained model used.
We exploit the characteristics of the on-device model (i.e., it is accessible but untrainable) to identify the pre-trained model and transfer learning approach for crafting a self-trained binary adversarial model similar to the targeted model to generate adversarial images.

\subsection{Security of Deep Learning Mobile Apps}
The security of mobile apps has become a major research field due to the ubiquity of smartphones \cite{amin2019androshield}.
Many studies \cite{merlo2017riskindroid,amin2019androshield} perform static or dynamic analysis or both to detect vulnerabilities in mobile apps.
However, there are few studies that focus on the security of deep learning mobile apps.
Wang et al. \cite{wang2018deep} and Xu et al. \cite{xu2019first} carried out empirical studies on challenges and current status of pushing deep learning towards mobile apps and found that most on-device models in mobile apps are exposed without any protection, which may result in severe security and privacy issues.
Although these studies reveal the security issues of deep learning mobile apps, they do not perform practical attacks to on-device models.
In contrast, our work represents a solid step towards the security of deep learning mobile apps (i.e., attacking on-device models).

Following the previous work \cite{xu2019first}, Sun et al. \cite{sun2021mind} analyzed the model protection problem of on-device machine learning models in mobile apps, and demonstrated the feasibility of stealing private models from AI apps as well as the potential financial losses that can occur.
Recently, Huang et al. \cite{huang2021robustness} investigated the vulnerability of deep learning models within real-world Android apps against adversarial attacks. 
Their findings unveiled that on-device deep learning models, which adopt or fine-tune pre-trained models, are more vulnerable to adversarial attacks.
However, these approaches either focus on the model extraction or the exploration of vulnerable models, they do not provide a practical way of attacking on-device models.
Hence, our work aims to fill this knowledge gap by proposing an effective model-specific adversarial attack which can be directly used for attacking practical deep learning apps.

\section{Conclusion}
\label{sec:conclusion}

This paper proposes a simple yet effective adversarial attack framework against on-device deep learning models.
Experiments from four perspectives, including pre-trained models, datasets, transfer learning approaches and adversarial attack algorithms, show that the attack is universal and robust.
Exemplifying three real-world deep learning mobile apps of skin cancer, traffic sign, and cash recognition demonstrate the potential damage of the attack.
We hope this work can raise the awareness of the community for protecting on-device models.
A set of avenues for future research include: 
First, in this paper, we only concentrate on attacking on-device models in Android apps. 
Considering the model accessibility in Android, it is also possible to transfer such attacks from Android to iOS or even to counterparts from other platforms like web, resulting in more powerful and generic attacks.
Second, this paper only considers attacks on deep learning models in smartphone, it is interesting to explore such attacks against models in other edge devices like Raspberry Pi.
Third, our study focuses on deep learning models related to image tasks, other task domains like natural language processing are also worth studying.
Finally, the countermeasures discussed in Section \ref{discussion:defence} may serve as a point of departure for developing effective defences.

\bibliographystyle{IEEEtran}
\bibliography{bibliography}

\begin{thebibliography}{10}
\providecommand{\url}[1]{#1}
\csname url@samestyle\endcsname
\providecommand{\newblock}{\relax}
\providecommand{\bibinfo}[2]{#2}
\providecommand{\BIBentrySTDinterwordspacing}{\spaceskip=0pt\relax}
\providecommand{\BIBentryALTinterwordstretchfactor}{4}
\providecommand{\BIBentryALTinterwordspacing}{\spaceskip=\fontdimen2\font plus
\BIBentryALTinterwordstretchfactor\fontdimen3\font minus
  \fontdimen4\font\relax}
\providecommand{\BIBforeignlanguage}[2]{{%
\expandafter\ifx\csname l@#1\endcsname\relax
\typeout{** WARNING: IEEEtran.bst: No hyphenation pattern has been}%
\typeout{** loaded for the language `#1'. Using the pattern for}%
\typeout{** the default language instead.}%
\else
\language=\csname l@#1\endcsname
\fi
#2}}
\providecommand{\BIBdecl}{\relax}
\BIBdecl

\bibitem{huang2021robustness}
Y.~Huang, H.~Hu, and C.~Chen, ``Robustness of on-device models: Adversarial
  attack to deep learning models on android apps,'' in \emph{IEEE/ACM 43rd
  International Conference on Software Engineering: Software Engineering in
  Practice (ICSE-SEIP)}, 2021.

\bibitem{sun2021mind}
Z.~Sun, R.~Sun, L.~Lu, and A.~Mislove, ``Mind your weight (s): A large-scale
  study on insufficient machine learning model protection in mobile apps,'' in
  \emph{30th $\{$USENIX$\}$ Security Symposium ($\{$USENIX$\}$ Security 21)},
  2021.

\bibitem{xu2019first}
M.~Xu, J.~Liu, Y.~Liu, F.~X. Lin, Y.~Liu, and X.~Liu, ``A first look at deep
  learning apps on smartphones,'' in \emph{The World Wide Web Conference},
  2019, pp. 2125--2136.

\bibitem{tensorflowlite}
``Tensorflow lite,'' \url{https://https://www.tensorflow.org/lite}, 2021.

\bibitem{pytorchmobile}
``Pytorch mobile,'' \url{https://pytorch.org/mobile/home/}, 2021.

\bibitem{goodfellow2014explaining}
I.~J. Goodfellow, J.~Shlens, and C.~Szegedy, ``Explaining and harnessing
  adversarial examples,'' \emph{arXiv preprint arXiv:1412.6572}, 2014.

\bibitem{suya2020hybrid}
F.~Suya, J.~Chi, D.~Evans, and Y.~Tian, ``Hybrid batch attacks: Finding
  black-box adversarial examples with limited queries,'' in \emph{29th USENIX
  Security Symposium USENIX Security 20)}, 2020, pp. 1327--1344.

\bibitem{chen2017zoo}
P.-Y. Chen, H.~Zhang, Y.~Sharma, J.~Yi, and C.-J. Hsieh, ``Zoo: Zeroth order
  optimization based black-box attacks to deep neural networks without training
  substitute models,'' in \emph{Proceedings of the 10th ACM workshop on
  artificial intelligence and security}, 2017, pp. 15--26.

\bibitem{chen2020hopskipjumpattack}
J.~Chen, M.~I. Jordan, and M.~J. Wainwright, ``Hopskipjumpattack: A
  query-efficient decision-based attack,'' in \emph{2020 ieee symposium on
  security and privacy (sp)}.\hskip 1em plus 0.5em minus 0.4em\relax IEEE,
  2020, pp. 1277--1294.

\bibitem{co2019procedural}
K.~T. Co, L.~Mu{\~n}oz-Gonz{\'a}lez, S.~de~Maupeou, and E.~C. Lupu,
  ``Procedural noise adversarial examples for black-box attacks on deep
  convolutional networks,'' in \emph{Proceedings of the 2019 ACM SIGSAC
  Conference on Computer and Communications Security}, 2019, pp. 275--289.

\bibitem{li2019adversarial}
X.~Li, S.~Ji, M.~Han, J.~Ji, Z.~Ren, Y.~Liu, and C.~Wu, ``Adversarial examples
  versus cloud-based detectors: A black-box empirical study,'' \emph{IEEE
  Transactions on Dependable and Secure Computing}, 2019.

\bibitem{brendel2017decision}
W.~Brendel, J.~Rauber, and M.~Bethge, ``Decision-based adversarial attacks:
  Reliable attacks against black-box machine learning models,'' \emph{arXiv
  preprint arXiv:1712.04248}, 2017.

\bibitem{li2021deeppayload}
Y.~Li, J.~Hua, H.~Wang, C.~Chen, and Y.~Liu, ``Deeppayload: Black-box backdoor
  attack on deep learning models through neural payload injection,'' in
  \emph{2021 IEEE/ACM 43rd International Conference on Software Engineering
  (ICSE)}.\hskip 1em plus 0.5em minus 0.4em\relax IEEE, 2021, pp. 263--274.

\bibitem{wang2020edge}
X.~Wang, Y.~Han, V.~C. Leung, D.~Niyato, X.~Yan, and X.~Chen, ``Edge computing
  for artificial intelligence,'' in \emph{Edge AI}.\hskip 1em plus 0.5em minus
  0.4em\relax Springer, 2020, pp. 97--115.

\bibitem{TensorHub}
Google, ``Tensorflow hub,'' \url{https://tfhub.dev/}, 2021.

\bibitem{modelzoo}
J.~Y. Koh, ``Model zoo,'' \url{https://modelzoo.co/}, 2021.

\bibitem{pytorch}
``Pytorch,'' \url{https://pytorch.org}, 2021.

\bibitem{coreml}
``Core ml,'' \url{https://developer.apple.com/documentation/coreml}, 2021.

\bibitem{ccurukouglu2018deep}
N.~{\c{C}}{\"u}r{\"u}ko{\u{g}}lu and B.~M. {\"O}zyildirim, ``Deep learning on
  mobile systems,'' in \emph{2018 Innovations in Intelligent Systems and
  Applications Conference (ASYU)}.\hskip 1em plus 0.5em minus 0.4em\relax IEEE,
  2018, pp. 1--4.

\bibitem{carlini2017towards}
N.~Carlini and D.~Wagner, ``Towards evaluating the robustness of neural
  networks,'' in \emph{2017 ieee symposium on security and privacy (sp)}.\hskip
  1em plus 0.5em minus 0.4em\relax IEEE, 2017, pp. 39--57.

\bibitem{rauber2020fast}
J.~Rauber and M.~Bethge, ``Fast differentiable clipping-aware normalization and
  rescaling,'' \emph{arXiv preprint arXiv:2007.07677}, 2020.

\bibitem{ilyas2018black}
A.~Ilyas, L.~Engstrom, A.~Athalye, and J.~Lin, ``Black-box adversarial attacks
  with limited queries and information,'' in \emph{International Conference on
  Machine Learning}.\hskip 1em plus 0.5em minus 0.4em\relax PMLR, 2018, pp.
  2137--2146.

\bibitem{papernot2017practical}
N.~Papernot, P.~McDaniel, I.~Goodfellow, S.~Jha, Z.~B. Celik, and A.~Swami,
  ``Practical black-box attacks against machine learning,'' in
  \emph{Proceedings of the 2017 ACM on Asia conference on computer and
  communications security}, 2017, pp. 506--519.

\bibitem{papernot2016transferability}
N.~Papernot, P.~McDaniel, and I.~Goodfellow, ``Transferability in machine
  learning: from phenomena to black-box attacks using adversarial samples,''
  \emph{arXiv preprint arXiv:1605.07277}, 2016.

\bibitem{guo2019simple}
C.~Guo, J.~Gardner, Y.~You, A.~G. Wilson, and K.~Weinberger, ``Simple black-box
  adversarial attacks,'' in \emph{International Conference on Machine
  Learning}.\hskip 1em plus 0.5em minus 0.4em\relax PMLR, 2019, pp. 2484--2493.

\bibitem{andriushchenko2020square}
M.~Andriushchenko, F.~Croce, N.~Flammarion, and M.~Hein, ``Square attack: a
  query-efficient black-box adversarial attack via random search,'' in
  \emph{European Conference on Computer Vision}.\hskip 1em plus 0.5em minus
  0.4em\relax Springer, 2020, pp. 484--501.

\bibitem{winsniewski2012apktool}
R.~Winsniewski, ``Apktool: a tool for reverse engineering android apk files,''
  \emph{https://ibotpeaches.github.io/Apktool/}, 2012.

\bibitem{levenshtein1966binary}
V.~I. Levenshtein, ``Binary codes capable of correcting deletions, insertions,
  and reversals,'' in \emph{Soviet physics doklady}, vol.~10, no.~8.\hskip 1em
  plus 0.5em minus 0.4em\relax Soviet Union, 1966, pp. 707--710.

\bibitem{netron}
L.~Roeder, ``Netron,'' \url{https://netron.app/}, 2021.

\bibitem{tensorflowtra}
``Tensorflow transfer learning,''
  \url{https://www.tensorflow.org/tutorials/images/transfer_learning}, 2021.

\bibitem{howard2017mobilenets}
A.~G. Howard, M.~Zhu, B.~Chen, D.~Kalenichenko, W.~Wang, T.~Weyand,
  M.~Andreetto, and H.~Adam, ``Mobilenets: Efficient convolutional neural
  networks for mobile vision applications,'' \emph{arXiv preprint
  arXiv:1704.04861}, 2017.

\bibitem{szegedy2016rethinking}
C.~Szegedy, V.~Vanhoucke, S.~Ioffe, J.~Shlens, and Z.~Wojna, ``Rethinking the
  inception architecture for computer vision,'' in \emph{Proceedings of the
  IEEE conference on computer vision and pattern recognition}, 2016, pp.
  2818--2826.

\bibitem{he2016deep}
K.~He, X.~Zhang, S.~Ren, and J.~Sun, ``Deep residual learning for image
  recognition,'' in \emph{Proceedings of the IEEE conference on computer vision
  and pattern recognition}, 2016, pp. 770--778.

\bibitem{perez2017effectiveness}
L.~Perez and J.~Wang, ``The effectiveness of data augmentation in image
  classification using deep learning,'' \emph{arXiv preprint arXiv:1712.04621},
  2017.

\bibitem{sandler2018mobilenetv2}
M.~Sandler, A.~Howard, M.~Zhu, A.~Zhmoginov, and L.-C. Chen, ``Mobilenetv2:
  Inverted residuals and linear bottlenecks,'' in \emph{Proceedings of the IEEE
  conference on computer vision and pattern recognition}, 2018, pp. 4510--4520.

\bibitem{he2016identity}
K.~He, X.~Zhang, S.~Ren, and J.~Sun, ``Identity mappings in deep residual
  networks,'' in \emph{European conference on computer vision}.\hskip 1em plus
  0.5em minus 0.4em\relax Springer, 2016, pp. 630--645.

\bibitem{deng2009imagenet}
J.~Deng, W.~Dong, R.~Socher, L.-J. Li, K.~Li, and L.~Fei-Fei, ``Imagenet: A
  large-scale hierarchical image database,'' in \emph{2009 IEEE conference on
  computer vision and pattern recognition}.\hskip 1em plus 0.5em minus
  0.4em\relax Ieee, 2009, pp. 248--255.

\bibitem{wang2018great}
B.~Wang, Y.~Yao, B.~Viswanath, H.~Zheng, and B.~Y. Zhao, ``With great training
  comes great vulnerability: Practical attacks against transfer learning,'' in
  \emph{27th $\{$USENIX$\}$ Security Symposium ($\{$USENIX$\}$ Security 18)},
  2018, pp. 1281--1297.

\bibitem{abdelkader2020headless}
A.~Abdelkader, M.~J. Curry, L.~Fowl, T.~Goldstein, A.~Schwarzschild, M.~Shu,
  C.~Studer, and C.~Zhu, ``Headless horseman: Adversarial attacks on transfer
  learning models,'' in \emph{ICASSP 2020-2020 IEEE International Conference on
  Acoustics, Speech and Signal Processing (ICASSP)}.\hskip 1em plus 0.5em minus
  0.4em\relax IEEE, 2020, pp. 3087--3091.

\bibitem{krizhevsky2009learning}
A.~Krizhevsky, G.~Hinton \emph{et~al.}, ``Learning multiple layers of features
  from tiny images,'' 2009.

\bibitem{Stallkamp-IJCNN-2011}
J.~Stallkamp, M.~Schlipsing, J.~Salmen, and C.~Igel, ``The {G}erman {T}raffic
  {S}ign {R}ecognition {B}enchmark: A multi-class classification competition,''
  in \emph{IEEE International Joint Conference on Neural Networks}, 2011, pp.
  1453--1460.

\bibitem{Nilsback08}
M.-E. Nilsback and A.~Zisserman, ``Automated flower classification over a large
  number of classes,'' in \emph{Proceedings of the Indian Conference on
  Computer Vision, Graphics and Image Processing}, Dec 2008.

\bibitem{ye2020detection}
D.~Ye, C.~Chen, C.~Liu, H.~Wang, and S.~Jiang, ``Detection defense against
  adversarial attacks with saliency map,'' \emph{arXiv preprint
  arXiv:2009.02738}, 2020.

\bibitem{rusak2020simple}
E.~Rusak, L.~Schott, R.~S. Zimmermann, J.~Bitterwolf, O.~Bringmann, M.~Bethge,
  and W.~Brendel, ``A simple way to make neural networks robust against diverse
  image corruptions,'' in \emph{European Conference on Computer Vision}.\hskip
  1em plus 0.5em minus 0.4em\relax Springer, 2020, pp. 53--69.

\bibitem{rauber2017foolbox}
J.~Rauber, W.~Brendel, and M.~Bethge, ``Foolbox: A python toolbox to benchmark
  the robustness of machine learning models,'' \emph{arXiv preprint
  arXiv:1707.04131}, 2017.

\bibitem{zhang2020two}
Y.~Zhang, Y.~Song, J.~Liang, K.~Bai, and Q.~Yang, ``Two sides of the same coin:
  White-box and black-box attacks for transfer learning,'' in \emph{Proceedings
  of the 26th ACM SIGKDD International Conference on Knowledge Discovery \&
  Data Mining}, 2020, pp. 2989--2997.

\bibitem{tentransfer}
``Tensorflow: Transfer learning and fine-tuning,''
  \url{https://www.tensorflow.org/tutorials/images/transfer\_learning}, 2021.

\bibitem{googlemobilevision}
``Google mobile vision,'' \url{https://developers.google.com/vision/}, 2021.

\bibitem{nasr2016melanoma}
E.~Nasr-Esfahani, S.~Samavi, N.~Karimi, S.~M.~R. Soroushmehr, M.~H. Jafari,
  K.~Ward, and K.~Najarian, ``Melanoma detection by analysis of clinical images
  using convolutional neural network,'' in \emph{2016 38th Annual International
  Conference of the IEEE Engineering in Medicine and Biology Society
  (EMBC)}.\hskip 1em plus 0.5em minus 0.4em\relax IEEE, 2016, pp. 1373--1376.

\bibitem{tan2019intelligent}
T.~Y. Tan, L.~Zhang, and C.~P. Lim, ``Intelligent skin cancer diagnosis using
  improved particle swarm optimization and deep learning models,''
  \emph{Applied Soft Computing}, vol.~84, p. 105725, 2019.

\bibitem{nahata2020deep}
H.~Nahata and S.~P. Singh, ``Deep learning solutions for skin cancer detection
  and diagnosis,'' in \emph{Machine Learning with Health Care
  Perspective}.\hskip 1em plus 0.5em minus 0.4em\relax Springer, 2020, pp.
  159--182.

\bibitem{grigorescu2020survey}
S.~Grigorescu, B.~Trasnea, T.~Cocias, and G.~Macesanu, ``A survey of deep
  learning techniques for autonomous driving,'' \emph{Journal of Field
  Robotics}, vol.~37, no.~3, pp. 362--386, 2020.

\bibitem{vadlamudidriver}
S.~Vadlamudi, ``Driver drowsiness detection on raspberry pi 4.''

\bibitem{warden2019tinyml}
P.~Warden and D.~Situnayake, \emph{Tinyml: Machine learning with tensorflow
  lite on arduino and ultra-low-power microcontrollers}.\hskip 1em plus 0.5em
  minus 0.4em\relax " O'Reilly Media, Inc.", 2019.

\bibitem{tfliteembedded}
``Tensorflow lite for embedded linux,''
  \url{https://www.tensorflow.org/lite/guide/python}, 2021.

\bibitem{tflitemicrocontrollers}
``Tensorflow lite for microcontrollers,''
  \url{https://www.tensorflow.org/lite/microcontrollers}, 2021.

\bibitem{tftrusted}
``Tf trusted,'' \url{https://github.com/capeprivacy/tf-trusted}, 2021.

\bibitem{niu2021decade}
S.~Niu, Y.~Liu, J.~Wang, and H.~Song, ``A decade survey of transfer learning
  (2010-2020),'' \emph{IEEE Transactions on Artificial Intelligence}, 2021.

\bibitem{ji2018model}
Y.~Ji, X.~Zhang, S.~Ji, X.~Luo, and T.~Wang, ``Model-reuse attacks on deep
  learning systems,'' in \emph{Proceedings of the 2018 ACM SIGSAC Conference on
  Computer and Communications Security}, 2018, pp. 349--363.

\bibitem{prabhu2018grey}
V.~U. Prabhu and J.~Whaley, ``On grey-box adversarial attacks and transfer
  learning,'' \emph{online: https://unify. id/wpcontent/uploads/2018/03/greybox
  attack.pdf}, 2018.

\bibitem{rezaei2019target}
S.~Rezaei and X.~Liu, ``A target-agnostic attack on deep models: Exploiting
  security vulnerabilities of transfer learning,'' \emph{arXiv preprint
  arXiv:1904.04334}, 2019.

\bibitem{pal2020transfer}
B.~Pal and S.~Tople, ``To transfer or not to transfer: Misclassification
  attacks against transfer learned text classifiers,'' \emph{arXiv preprint
  arXiv:2001.02438}, 2020.

\bibitem{amin2019androshield}
A.~Amin, A.~Eldessouki, M.~T. Magdy, N.~Abdeen, H.~Hindy, and I.~Hegazy,
  ``Androshield: automated android applications vulnerability detection, a
  hybrid static and dynamic analysis approach,'' \emph{Information}, vol.~10,
  no.~10, p. 326, 2019.

\bibitem{merlo2017riskindroid}
A.~Merlo and G.~C. Georgiu, ``Riskindroid: Machine learning-based risk analysis
  on android,'' in \emph{Ifip international conference on ict systems security
  and privacy protection}.\hskip 1em plus 0.5em minus 0.4em\relax Springer,
  2017, pp. 538--552.

\bibitem{wang2018deep}
J.~Wang, B.~Cao, P.~Yu, L.~Sun, W.~Bao, and X.~Zhu, ``Deep learning towards
  mobile applications,'' in \emph{2018 IEEE 38th International Conference on
  Distributed Computing Systems (ICDCS)}.\hskip 1em plus 0.5em minus
  0.4em\relax IEEE, 2018, pp. 1385--1393.

\end{thebibliography}

\vspace{-10cm}
\begin{IEEEbiography}[{\includegraphics[width=1in,height=1.25in,clip,keepaspectratio]{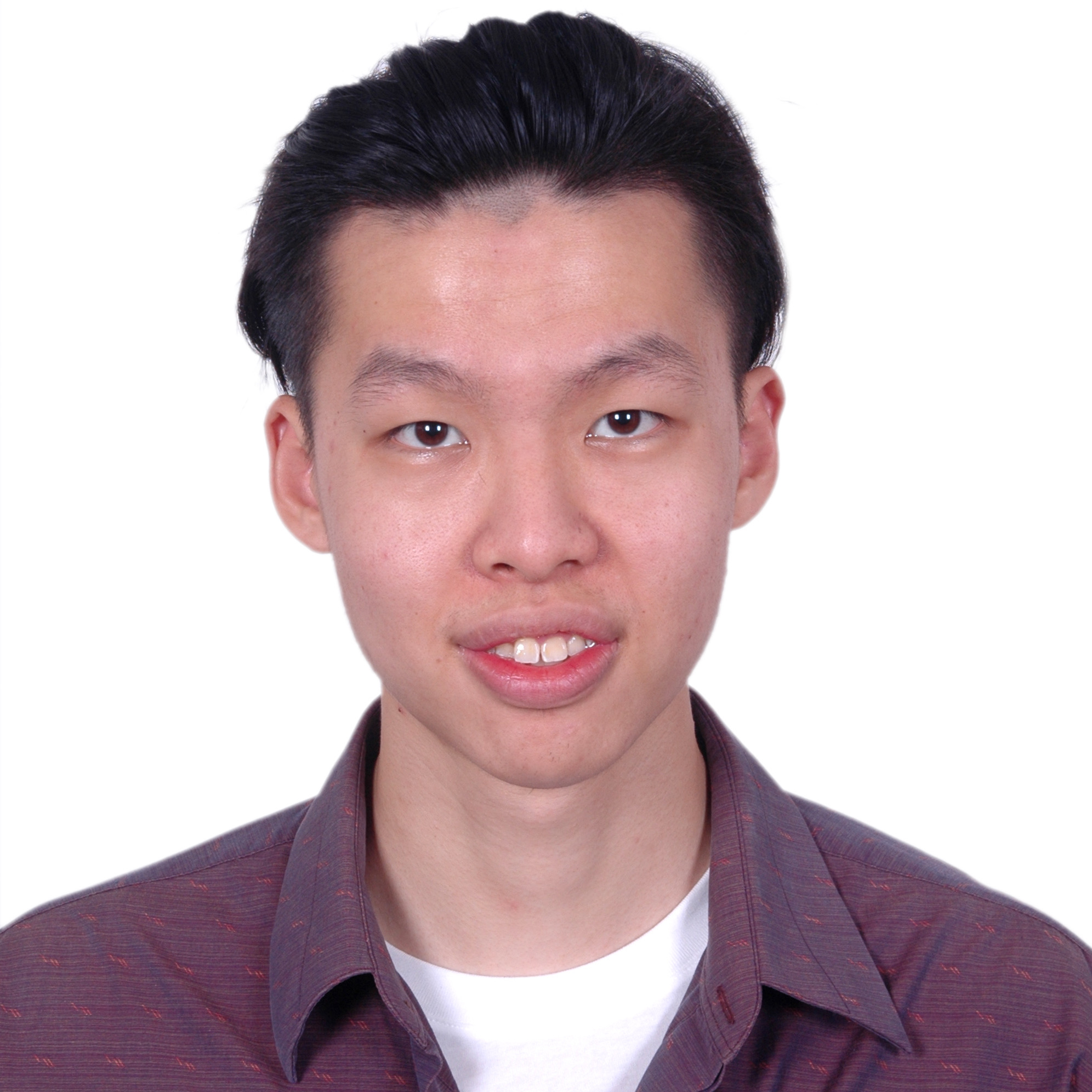}}]{Yujin Huang} is currently pursuing the Ph.D. degree with the
Faculty of Information Technology, Monash University, Australia. 
His research concentrates on causal discovery, natural language processing, and deep learning security.
\end{IEEEbiography}

\vspace{-10cm}
\begin{IEEEbiography}[{\includegraphics[width=1in,height=1.25in,clip,keepaspectratio]{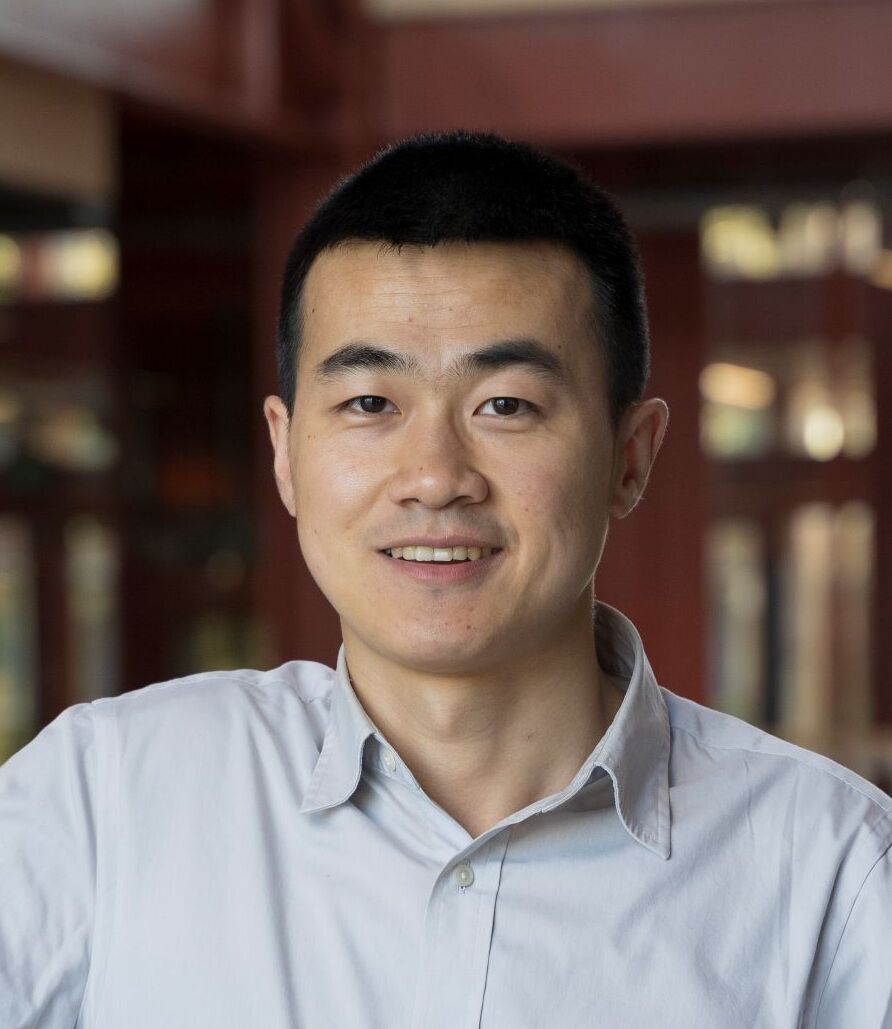}}]{Chunyang Chen} is a lecturer (Assistant Professor) in Faculty of Information Technology, Monash University, Australia. 
His research focuses on software engineering, deep learning
and human-computer interaction. 
He has published over 40 papers in referred journals or
conferences. 
He is a member of IEEE and ACM.
He has received ACM SIGSOFT Distinguished Paper Award in ICSE 2020, Facebook Research Award in Probability and Programming 2020, etc. https://chunyang-chen.github.io/
\end{IEEEbiography}







\end{document}